\documentclass[11pt]{article}

\usepackage[utf8]{inputenc}
\usepackage[T1]{fontenc}
\usepackage{lmodern}                       
\usepackage{microtype}                     
\usepackage{amsmath,amssymb}
\usepackage{graphicx}
\usepackage{booktabs}
\usepackage[colorlinks=true,linkcolor=black,citecolor={blue!50!black},urlcolor={blue!50!black}]{hyperref}
\usepackage{url}
\usepackage{natbib}
\usepackage{xcolor}
\usepackage{caption}
\usepackage[hmargin=2.5cm,vmargin=2.5cm]{geometry}

\usepackage{titling}
\pretitle{\par\centering\Large\bfseries}
\posttitle{\par\vskip 0.6em \noindent\rule{\textwidth}{0.4pt}\par\vskip 0.5em}
\preauthor{\par\centering\normalsize}
\postauthor{\par}
\predate{}\postdate{}

\newcommand{\cmark}{\textcolor{green!60!black}{\ding{51}}}
\newcommand{\xmark}{\textcolor{red!70}{\ding{55}}}
\newcommand{\pmark}{\textcolor{orange!80!black}{$\sim$}}
\usepackage{pifont}

\title{\textbf{FundaPod: A Multi-Persona Agent Pod Architecture\\with Knowledge Graph Memory for AI-Assisted\\Fundamental Investment Research}}

\author{%
\begin{tabular}{c@{\hspace{2.5em}}c@{\hspace{2.5em}}c}
Di Zhu\thanks{Corresponding author.} & Lei (Nico) Zheng & Zihan Chen \\
Stevens Institute of Technology      & UMass Boston      & Stevens Institute of Technology \\
Hoboken, NJ, USA                     & Boston, MA, USA   & Hoboken, NJ, USA \\
\texttt{dzhu1@stevens.edu}           & \texttt{lei.zheng@umb.edu} & \texttt{zchen61@stevens.edu}
\end{tabular}%
}

\date{}

\begin{document}

\maketitle

\begin{abstract}
Large language models (LLMs) are increasingly applied in finance, yet most existing work emphasizes trading signals or financial NLP tasks centered on prediction. Institutional fundamental research, by contrast, requires human analysts or AI agents to gather evidence, identify business drivers, compare competing viewpoints, and generate investment memos. Its broader goal is not merely to predict outcomes, but to produce investment plans that are transparent, reusable, and verifiable, while contributing to the cumulative development of investment knowledge. We present \textbf{FundaPod}, a multi-persona agent pod architecture for AI-assisted fundamental investment research. We argue that fundamental research is a \emph{human-centric decision-support task} that is qualitatively distinct from trading-signal generation, and is therefore better served by an independence-preserving architecture. In FundaPod, AI agents with different personas, such as value investors or macro strategists, conduct research independently under a shared provenance contract. Their disagreements are then surfaced post hoc for adjudication by the human portfolio manager (PM) through a knowledge-graph memory system. This paper contributes five design principles for human-AI hybrid systems supporting fundamental research, grounded in design-science practice and theories of cognitive isolation and human-machine coordination. It also describes four architectural mechanisms: a persona distillation pipeline that turns public investor materials into deployable agents; a declarative skill registry that lets the planner derive typed task graphs; a grounded evidence model that links memo claims to verifiable sources; and a knowledge-graph “second brain” that connects tickers, memos, analysts, and themes. We demonstrate the architecture through a complete case study and a persona-based memo comparison.
\end{abstract}

\section{Introduction}
\label{sec:intro}

Fundamental investment research at institutional firms follows a structured and repeatable process. Analysts begin by initiating or updating coverage on a company. They review public filings, market data, news, and other relevant sources. They then extract key performance indicators (KPIs) for the target company, identify business drivers, and assess how those drivers affect the investment case. Afterward, they synthesize the evidence into research memos, investment theses, and monitoring plans for the coverage universe. This work requires financial judgment, careful evidence handling, and the ability to connect information across companies, sectors, and macro themes. These requirements make fundamental research a valuable but difficult setting for LLM-based automation.

Most existing applications of LLMs in finance address only part of this workflow. Domain-specific financial LLMs improve performance on financial language tasks~\citep{wu2023bloomberggpt, yang2025fingptopensourcefinanciallarge,Chen2023ChatGPT}, but they are not designed to manage an end-to-end investment research process. LLM-based trading agents use market data, memory, and specialized roles to generate investment or trading decisions~\citep{zhang2024finagent, yu2025finmem, xiao2025tradingagentsmultiagentsllmfinancial}. These systems are useful for studying trading behavior, but their primary objective is signal generation rather than research production. Fundamental research has a different goal. It must produce a reasoned view that can be inspected, challenged, updated, and reused over time.

General multi-agent systems provide another important building block. Frameworks such as AutoGen~\citep{wu2024autogen}, MetaGPT~\citep{hong2023metagpt}, ChatDev~\citep{qian2023chatdev}, and Manus~\citep{manus2025} show that role-based agents can coordinate to facilitate complex tasks. However, these systems are usually designed for broad productivity or software engineering workflows. They do not directly address the needs of fundamental investment research teams, where agents must combine financial expertise, source-level evidence, and persistent memory across many research engagements.

A related challenge is memory. Fundamental research is cumulative. A useful system should not only answer one question or write one memo. It should remember prior coverage, link new evidence to old conclusions, and surface connections across companies and themes. Knowledge graphs are well suited to this problem because they represent entities and relationships explicitly. Recent work on graph-based agent memory and financial knowledge graphs shows the value of this structure~\citep{anokhin2024arigraph, yang2026graphbasedagentmemorytaxonomy, magma2026, findkg2024}. However, existing systems do not use knowledge graphs as a shared memory for a team of investment agents.

In this paper, we present \textbf{FundaPod}, a multi-persona agent architecture for institutional fundamental investment research\footnote{\url{https://github.com/dgtql/FundaPod}}. Our central thesis is that fundamental research is a class of \emph{human-centric decision-support tasks} qualitatively distinct from trading-signal generation: it must produce evidence-backed views that a human portfolio manager (PM) can inspect, contest, and selectively adopt. We argue that this objective is best served not by debate-converging multi-agent systems, but by an \emph{independence-preserving} architecture in which persona-distilled agents reason in isolation under a common provenance contract, and disagreement is surfaced post-hoc for the human expert to adjudicate through a shared knowledge-graph memory. This positions FundaPod as a human-AI hybrid platform~\citep{rai2019NextGenerationDigital, seeber2020MachinesTeammatesResearch, berente2021ManagingArtificialIntelligence} designed to augment, rather than substitute for, the analyst's reasoning~\citep{raisch2021AutomationAugmentationParadox}.

We make two contributions in this paper:
\begin{enumerate}
    \item \textbf{A design-principle account of research-production decision support.} We articulate five design principles (\S\ref{sec:design_principles}) for human-AI hybrid systems supporting fundamental research, anchored in design-science research practice~\citep{hevner2004DesignScienceIS} and a kernel theory of cognitive isolation under informational cascades, productive delegation, and human-machine coordination.

    \item \textbf{FundaPod, an architectural instantiation of these principles.} We describe the architecture, components, and design rationale of a deployed system that realizes the five principles through four mechanisms: \emph{persona-distilled agents} that supply distinct analytical lenses for the same target; a \emph{declarative skill registry} that lets the planner derive typed task graphs from each skill's \texttt{needs}/\texttt{produces} contract; a \emph{grounded evidence model} that links every generated claim to an inspectable source artifact; and a \emph{knowledge-graph ``second brain''} that connects tickers, memos, analysts, evidence, and themes into a queryable record of the pod's cumulative coverage.
\end{enumerate}

The architecture separates \emph{deterministic skills} from \emph{agent skills}. Deterministic skills handle programmatic tasks such as data ingestion, KPI extraction, and structured transformations. Agent skills use LLMs for judgment-heavy tasks such as memo writing, thematic exploration, and thesis comparison. This separation keeps repeatable pipeline work reproducible while using language models where interpretation and synthesis matter most.

FundaPod is also designed to be data-source agnostic. SEC filings, market data APIs, news feeds, alternative-data providers, and proprietary internal databases can all be connected through the same extensible skill interface. This makes the system adaptable to different investment teams, data stacks, and research processes.

The remainder of this paper is organized as follows. Section~\ref{sec:related} reviews related work on financial LLM agents, multi-agent frameworks, knowledge graphs, and agent memory. Section~\ref{sec:design_principles} articulates the five design principles that motivate the system. Section~\ref{sec:architecture} presents the system architecture. Section~\ref{sec:components} describes the core components in detail. Section~\ref{sec:case_study} walks through an illustrative case study and demonstrates (Appendices B–C) how an externally authored persona pack changes the reasoning structure imposed on the same upstream evidence, making the independence-then-synthesis principle concrete in operation. Section~\ref{sec:discussion} discusses design trade-offs, extensibility, and limitations. Section~\ref{sec:conclusion} concludes with future work.

\section{Related Work}
\label{sec:related}

FundaPod builds on four related streams of work: financial LLMs, LLM-based trading agents, multi-agent systems, and graph-based memory. Each stream contributes an important building block. Financial LLMs show the value of domain adaptation. Trading agents show how LLMs can act inside market workflows. Multi-agent systems show how complex tasks can be divided across specialized roles. Knowledge graphs provide a structure for persistent and relational memory. FundaPod combines these ideas in a different setting: institutional fundamental research.

\subsection{LLM Agents in Finance}

Early financial LLM research focused on adapting language models to financial text ~\citep{Chen2023ChatGPT}. BloombergGPT ~\citep{wu2023bloomberggpt} trains a 50-billion parameter model on proprietary financial data and improves performance on financial NLP tasks. FinGPT~\citep{yang2025fingptopensourcefinanciallarge} offers an open-source alternative based on lightweight fine-tuning. These systems show that financial domain adaptation improves model performance. However, they remain primarily model-centric. They do not define an agent workflow for collecting evidence, assigning research tasks, producing investment memos, and maintaining coverage over time.

A second line of work places LLMs inside trading workflows. FinAgent~\citep{zhang2024finagent} introduces a multimodal foundation agent for trading, with reflection and memory retrieval modules. FinMem~\citep{yu2025finmem} uses layered memory inspired by human trader cognition, including sensory, short-term, and long-term memory. TradingAgents~\citep{xiao2025tradingagentsmultiagentsllmfinancial} organizes specialized roles such as fundamental analysts, sentiment analysts, technical analysts, and risk managers into a staged trading pipeline. More recently, \citet{miyazaki2026expertteams} decompose investment analysis into fine-grained tasks assigned to expert agent teams.

These systems demonstrate that financial agents can combine data, memory, and role specialization. They also show that LLMs can support decision workflows rather than only perform isolated NLP tasks. However, their primary objective is still trading performance. They are usually evaluated through signals, portfolio returns, or benchmark trading outcomes. Fundamental research has a different objective. It produces evidence-backed views that can be inspected, challenged, revised, and reused by an investment team.

Recent benchmarks reinforce this distinction. AI-Trader~\citep{aitrader2025} evaluates LLM trading agents in live markets across U.S. stocks, A-shares, and cryptocurrencies, and shows that general LLM ability does not automatically translate into trading ability. FINSABER~\citep{finsaber2026} studies LLM investing strategies over longer horizons and broader symbol universes. \citet{aiagentsfinancialmarkets2026} examine the systemic implications of AI agents in financial markets, including efficiency gains, third-party dependencies, and correlated model behavior. \citet{ding2024llmagentfinancialtrading} survey LLM trading agents and classify them into direct traders and alpha miners. FundaPod differs from this literature by treating research production, rather than trade generation, as the core system objective.

\subsection{Multi-Agent Architectures}

The broader multi-agent literature provides the coordination patterns needed for complex agent workflows. AutoGen~\citep{wu2024autogen} introduces conversable agents and conversation programming for flexible interaction among agents. MetaGPT~\citep{hong2023metagpt} encodes standard operating procedures into an assembly-line structure with well-defined intermediate outputs. ChatDev~\citep{qian2023chatdev} frames software development as a virtual company made up of specialized agents. Manus~\citep{manus2025} represents a newer class of general-purpose autonomous agents that orchestrate specialized sub-agents in sandboxed environments for real-world task completion.

This line of work shows that multi-agent systems can divide complex work into specialized roles. It also shows that coordination logic is central to system quality. A useful agent system is not just a collection of prompts. It needs state, tools, task assignment, and control over intermediate outputs. Production frameworks such as LangGraph, CrewAI, and OpenAI's Agents SDK now make role-based orchestration, state management, and tool integration easier to implement~\citep{luo2025llmagentsurvey,infante2026aiagents}. These frameworks provide general infrastructure, but they do not define the financial research workflow itself.

FundaPod extends this line of work through its declarative skill registry. Each skill states the data it requires and the outputs it produces. The planner can then construct task graphs from these declarations. This differs from workflows that require developers to manually wire each task sequence. For investment teams, this design is important because research workflows change frequently. Teams may add a data vendor, introduce a new KPI template, or expand into a new sector. The system should support these changes without requiring orchestration code to be rewritten.

FundaPod also differs from multi-agent systems that rely on debate, voting, or shared context among agents. In FundaPod, personas reason independently before their outputs are synthesized. This design reflects evidence that much of the benefit in multi-agent debate comes from ensembling independent outputs, while repeated debate can reinforce errors or create groupthink, especially when agents are similar~\citep{smit2024mad}. It also resembles the organizational logic of multi-manager pod shops such as Millennium, Citadel, and Point72, where independent teams pursue distinct sources of alpha under centralized oversight. Recent empirical work reports that this structure can reduce volatility while preserving exposure to manager-specific alpha~\citep{blotnick2025podshops}. The same concern appears in the informational cascade model of \citet{bikhchandani1992cascades}: once agents observe one another's actions, private information may stop being reflected in the aggregate signal.

FundaPod applies this independence principle to research rather than portfolio management. Each persona forms its view separately. Cross-persona synthesis occurs after individual outputs are written into the Knowledge Graph. The master agent can then compare views, identify disagreements, and surface coverage gaps without allowing one persona to influence another too early.

\subsection{Knowledge Graphs for Agent Memory and Finance}
\label{sec:related_kg}

As agent systems move beyond single-session tasks, memory becomes a core design problem. Long context windows are useful, but they do not replace structured memory. Agents still need to retrieve relevant past work, preserve relationships among entities, and distinguish durable knowledge from temporary observations. Knowledge graphs are well suited to this problem because they store information as entities and relationships rather than as isolated text chunks.

Recent work has shown the value of graph-based memory for LLM agents. AriGraph~\citep{anokhin2024arigraph} demonstrates that agents using knowledge graph world models with episodic memory can outperform agents that rely on unstructured memory. \citet{yang2026graphbasedagentmemorytaxonomy} provide a taxonomy of graph-based agent memory and identify graph structures as effective for representing relational dependencies and improving retrieval. MAGMA~\citep{magma2026} introduces a multi-graph architecture that separates associative proximity from causal structure, allowing agents to reason about why events occur rather than only retrieving what occurred. GAM~\citep{wu2026gamhierarchicalgraphbasedagentic} addresses memory contamination through hierarchical graph-based memory and write isolation. \citet{agentmemorysurvey2026} further argue that external memory remains important even when context windows exceed 200K tokens, because agents still need cross-session persistence and selective retrieval.

Graph-based retrieval is also important for grounded generation. GraphRAG methods~\citep{graphrag2025survey} use graph structure to retrieve relational evidence that flat vector search may miss. Recent work goes further by combining graph-based retrieval with agentic reasoning loops --- surveyed by~\citet{chen2026SurveyAgenticGraphRAG} as \emph{agentic GraphRAG} --- in which the graph becomes the space the agent plans, acts, critiques, and remembers over, rather than a passive lookup table. This matters in finance because an investment claim often depends on links among a company, a filing, a KPI, a peer group, a macro theme, and a prior analyst view. A flat retrieval system can find similar text. A graph can show how the evidence is connected.

Financial knowledge graphs apply similar ideas to market information. FinDKG~\citep{findkg2024} builds dynamic knowledge graphs from financial news and uses them to detect global market trends for thematic investing. This shows that knowledge graphs can support financial analysis when relationships among events, companies, and themes matter. However, FinDKG operates mainly at the news-article level. It does not organize the research outputs of multiple agents or preserve an investment team's evolving coverage memory.

FundaPod uses the knowledge graph as the shared research memory of the pod. The graph connects companies, filings, KPIs, memos, personas, themes, and evidence artifacts. This allows the system to surface cross-company patterns, compare persona views, and identify areas where coverage is thin.

\subsection{Persona-Conditioned Agents}

The final line of related work concerns persona-conditioned agents. Generative Agents~\citep{park2023generativeagents} show that persona-driven agents can produce believable behavior when they maintain memories, reflections, and plans. MemGPT~\citep{packer2023memgpt} introduces tiered memory management inspired by operating systems. Voyager~\citep{wang2023voyager} builds a growing library of executable skills for embodied agents, where skills are reusable and compositional.

These systems provide three design patterns that are relevant to FundaPod. Personas can shape behavior. Memory can support continuity over time. Reusable skills can make agents more capable than one-off prompts. FundaPod combines these patterns in the setting of fundamental investment research. Personas represent distinct investment methods. The Knowledge Graph stores persistent research memory across companies and sessions. The skill registry provides composable tools for data ingestion, KPI extraction, memo writing, and thematic exploration.

The key difference is financial grounding. FundaPod does not use personas only to make agents sound different. It uses persona conditioning to create distinct analytical lenses that resemble how investment teams divide research judgment. A value investor, a macro strategist, and a quantitative analyst may examine the same company but ask different questions. FundaPod preserves those differences, grounds their claims in evidence, and stores their outputs in a shared research memory for later synthesis.

\section{Design Principles for Human-Centric Research Production}
\label{sec:design_principles}

\begin{figure}[!htbp]
\centering
\includegraphics[width=0.98\textwidth]{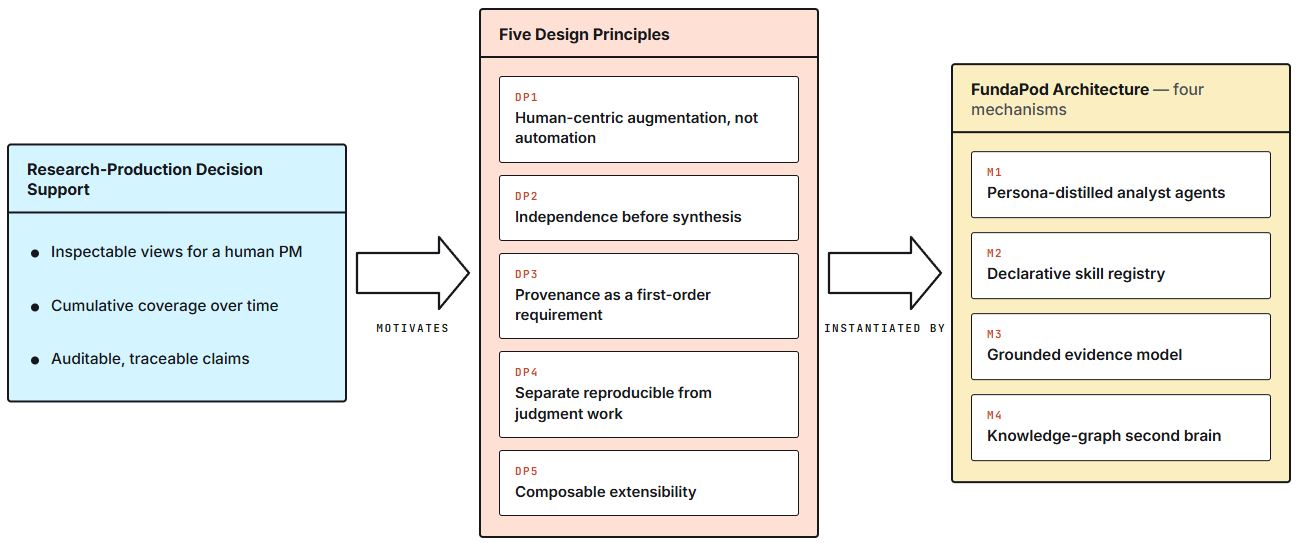}
\caption{%
\textbf{From problem to architecture: the FundaPod design rationale.}
Fundamental research production (left) is a class of human-centric decision-support tasks that places three requirements on a supporting system: inspectable views for a human PM, cumulative coverage over time, and auditable, traceable claims. These requirements motivate five \emph{design principles} (middle, DP1--DP5), which are in turn instantiated by FundaPod through four architectural mechanisms (right): persona-distilled agents, a declarative skill registry, a grounded evidence model, and a knowledge-graph ``second brain''.}
\label{fig:framework}
\end{figure}

Fundamental investment research is not signal generation; it is a class of decision-support tasks in which a human portfolio manager (PM) must form, defend, and revise an inspectable investment view over time. Following design-science research practice~\citep{hevner2004DesignScienceIS}, this section makes explicit the requirements such a task places on a system, and articulates five \emph{design principles} (DPs) that we used to shape FundaPod. The principles position the system as a \emph{human-AI hybrid}~\citep{rai2019NextGenerationDigital, seeber2020MachinesTeammatesResearch} whose role is to augment the PM's reasoning bandwidth rather than substitute for it. The architecture in \S\ref{sec:architecture} and the components in \S\ref{sec:components} are instantiations of these principles; the discussion in \S\ref{sec:discussion} revisits each principle as a deliberate trade-off against alternatives. Figure~\ref{fig:framework} summarizes the chain from the problem characterization through the five design principles to the four architectural mechanisms that realize them.

\paragraph{DP1: Human-centric augmentation, not automation.}
The system is designed to expand the PM's capacity for evidence gathering, lens application, and synthesis, while leaving the locus of judgment with the human expert. We treat the automation-augmentation choice as a normative commitment rather than a default outcome: an AI that imitates a human analyst is less valuable than one that augments the analyst's reasoning~\citep{raisch2021AutomationAugmentationParadox}. In FundaPod, this principle is realized through outputs designed to be inspected, contested, and selectively adopted: the master agent and Knowledge Graph view (\S\ref{sec:architecture}--\S\ref{sec:kg}) answer to the PM rather than to downstream automated action.

\paragraph{DP2: Independence before synthesis.}
Persona-distilled agents reason in isolation; synthesis is performed afterward, by the PM or by a master agent acting on the PM's behalf, not through inter-agent debate. Two bodies of work motivate this. Informational-cascade theory~\citep{bikhchandani1992cascades} warns that once agents observe one another's actions, private signals stop being reflected in the aggregate. Controlled studies of LLM multi-agent debate find that much of the apparent benefit comes from independent-output ensembling, while additional debate rounds can harden errors among similar agents~\citep{smit2024mad}. More broadly, IS scholarship on coordinating human and machine learning emphasizes that the structure of coordination, not the raw capability of either party, drives effective combined learning~\citep{sturm2021CoordinatingHumanMachine}. FundaPod operationalizes this in \S\ref{sec:architecture} L3: personas do not exchange context during reasoning, and the Knowledge Graph (\S\ref{sec:kg}) is read by the master agent and the PM, not by the agents themselves.

\paragraph{DP3: Provenance as a first-order requirement.}
Every claim produced by the system must be traceable to an inspectable source artifact. Critical-judgment settings have shown that professionals struggle to adopt AI outputs whose reasoning is opaque, and that ``engaged augmentation'' depends on practices that let the expert relate their own knowledge claims to the system's~\citep{lebovitz2022EngageOrNot}. The explainable-AI literature documents the cost of black-box behavior across decision-support domains~\citep{guidotti2019SurveyExplainingBlackBox, reinhard2026EffectsExplanationsXAI}. We therefore treat provenance not as a post-hoc explanation layer but as the storage substrate of the system: claims are linked to content-addressed evidence artifacts and the lineage is preserved across versions (\S\ref{sec:evidence}).

\paragraph{DP4: Separation of reproducible work from judgment work.}
Tasks that should be deterministic and auditable, such as data ingestion, KPI extraction, and source-quality gating, are implemented as deterministic or constrained-hybrid skills, while genuinely open-ended reasoning is reserved for agent skills under a shared contract. Empirical work on productive human-AI collaboration shows that effective delegation depends on each party doing what it can do reliably, and that humans benefit most when they can identify the boundary~\citep{fugener2022CognitiveChallengesHuman, lebovitz2022EngageOrNot}. In FundaPod, the Skill Registry (\S\ref{sec:registry}) exposes a single \texttt{needs}/\texttt{produces} contract behind three runner types (Deterministic, Hybrid, Agent), so the same pipeline can be reasoned about even where individual stages call language models.

\paragraph{DP5: Composable extensibility along independent axes.}
A decision-support system for evolving research workflows must absorb new data sources, new analytical skills, new personas, and new engagement workflows without forcing rewrites along the other axes. We follow the design-principles template advanced by~\citet{dellermann2019DesignPrinciplesHybrid} for hybrid-intelligence decision-support systems: declarative contracts at component boundaries so that extensions along one axis become immediately available to every existing user of the contract. In FundaPod, this is realized as four parallel extension points (data sources, skills, personas, workflows), each mediated by the same metadata contract (\S\ref{sec:discussion}).

\medskip
\noindent
Sections~\ref{sec:architecture}--\ref{sec:components} describe the architecture and components through which these principles are instantiated. The case study (\S\ref{sec:case_study}) walks the principles through a complete research engagement, and the discussion (\S\ref{sec:discussion}) revisits each as a deliberate trade-off against alternatives.

\section{System Architecture}
\label{sec:architecture}

FundaPod uses a \emph{pod} metaphor to model how fundamental research teams work. A pod contains multiple AI agents, each shaped by a different investor persona. These agents share the same infrastructure, but they approach a company through different analytical lenses. The goal is to produce research that is not only fluent, but also structured, traceable, and useful for institutional equity analysis.

\begin{figure}[!htbp]
\centering
\includegraphics[width=0.98\textwidth]{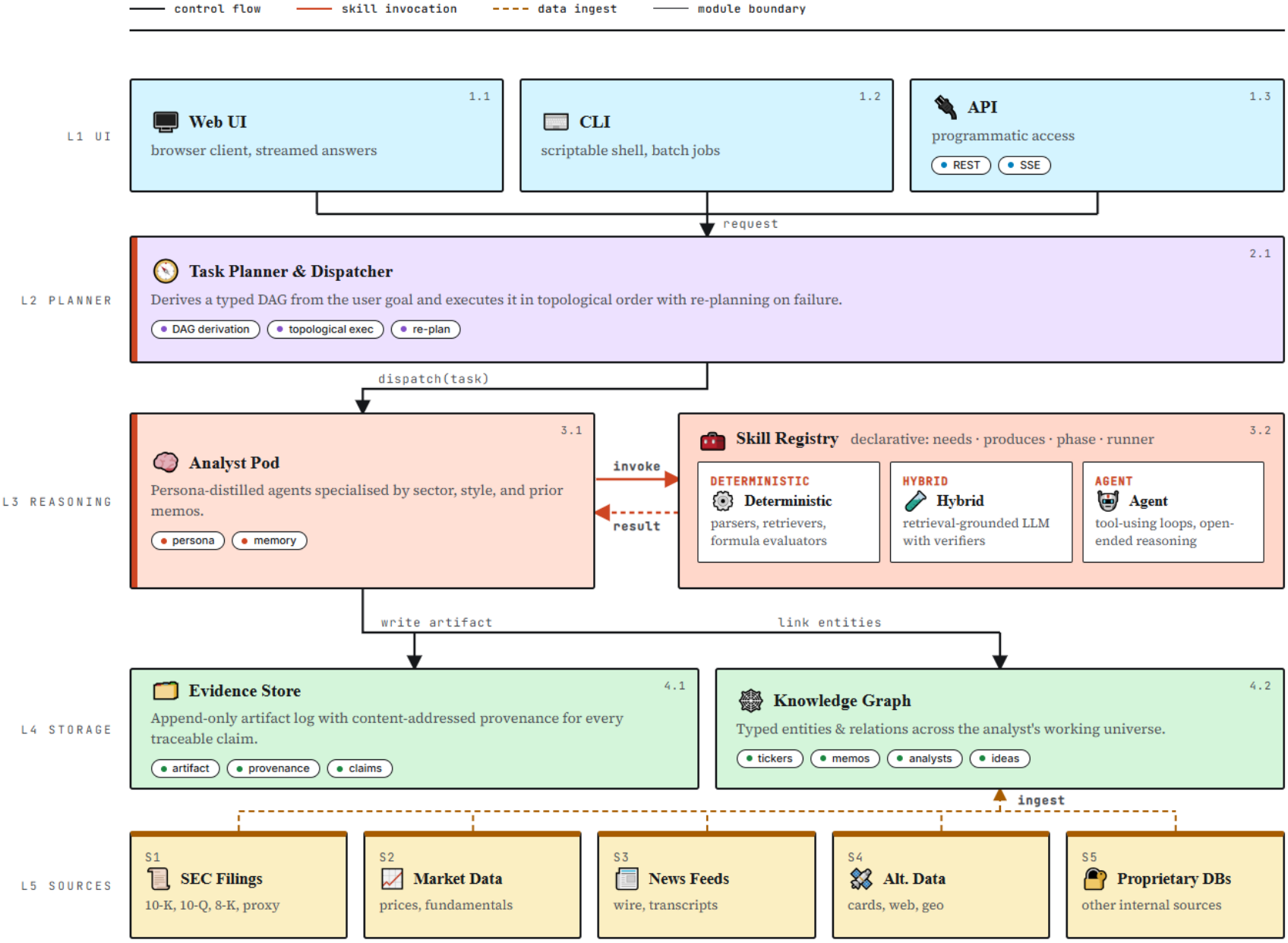}
\caption{%
\textbf{FundaPod system architecture.}
A request enters through the interface layer (\textsc{L1}) and is converted into a research plan by the Task Planner \& Dispatcher (\textsc{L2}).
The planner builds a typed DAG of skill invocations from the contracts declared in the Skill Registry.
Persona-distilled agents in the Analyst Pod (\textsc{L3}) use these skills to gather evidence, perform analysis, and produce research outputs.
Each skill output is written to the Evidence Store and linked into the Knowledge Graph (\textsc{L4}), so memo claims can be traced back to source data from filings, market data, news, transcripts, alternative data, or internal databases (\textsc{L5}).
Solid arrows show control flow, the blue arrow shows the Analyst--Skill invocation loop, and dashed arrows show data ingest from sources into storage.}
\label{fig:architecture}
\end{figure}

Our system is not implemented as one large agent or as an unstructured swarm of agents. Instead, it is organized into five horizontal layers, as shown in Figure~\ref{fig:architecture}. Each layer exposes a clear contract to the layers around it. This design keeps the main responsibilities separate. Persona-driven reasoning is handled by agents. Deterministic data work is handled by skills and pipelines. Evidence is stored in an append-only form so that later claims can be traced back to their sources.

This separation is important for reliability. In a research setting, a bad synthesis should not corrupt the source data. A failed data ingestion step should not change a persona. A new memo workflow should not require rebuilding the storage layer. By separating these concerns, FundaPod makes the research process easier to inspect, debug, and extend.

The layered design also gives the system four independent extension points. New data sources can be added at L5. New research skills can be added through the Skill Registry at L3. New personas can be added to the Pod at L3. New engagement workflows can be added as plan templates at L2. Each extension point can change without forcing changes to the others, which makes the system adaptable to different research teams, asset classes, and data environments (\S\ref{sec:discussion}).

\paragraph{\textsc{L1} -- Interface (Web UI, CLI, API).}
The interface layer gives users several ways to start the same research workflow. A portfolio manager may launch an engagement through the web interface, a developer may call the API, and an investment team may schedule a refresh through the command line. All three entry points are converted into the same internal request format before they reach the planner. As a result, the downstream research pipeline does not depend on how the request was submitted.

\paragraph{\textsc{L2} -- Task Planner \& Dispatcher.}
The task planner turns a user goal into an executable research plan. It reads the goal together with the metadata advertised by each skill in the registry, then builds a typed directed acyclic graph (DAG) of skill invocations. The edges in this graph represent declared data dependencies, not hard-coded function calls. The dispatcher then executes the graph in dependency order. If a step fails, the system can retry or re-plan instead of abandoning the full engagement. Because the plan is built from declared skill contracts, a newly added producer skill becomes available to downstream consumers without changes to the planner (\S\ref{sec:registry}).

\paragraph{\textsc{L3} -- Reasoning (Analyst Pod \& Skill Registry).}
The reasoning layer is where investment judgment enters the system. The Analyst Pod contains persona-distilled agents (\S\ref{sec:persona}), each with its own analytical style and memory of prior memos and coverage. A key design choice is that agents do not talk to one another and do not share memory during their core reasoning. Each persona works only from its own prior outputs and from artifacts retrieved through the Evidence Store.

This design mirrors the pod model used by multi-manager hedge funds. In that model, independent portfolio managers run separate books with limited coordination, so the overall fund can benefit from diverse sources of risk and return~\citep{blotnick2025podshops}. The LLM literature points to a similar concern. Shared context and debate can sometimes reinforce early errors, while much of the benefit from multi-agent protocols can come from combining independent outputs~\citep{smit2024mad}.

A \emph{master agent} sits above the pod and is the only component with visibility across agents. It uses the Knowledge Graph (\S\ref{sec:kg}) to answer questions that span multiple personas, compare views, and identify gaps in coverage. However, it does not feed one agent's reasoning into another agent's context. This preserves independent judgment before synthesis.

Analysts access system capabilities through the Skill Registry. The registry separates a skill's \emph{contract}, including its \texttt{needs}, \texttt{produces}, phase, and runner metadata, from its implementation. FundaPod supports three runner types under this shared contract. \textsc{Deterministic} skills handle reproducible tasks such as parsing, retrieval, and formula evaluation. \textsc{Hybrid} skills combine retrieval-grounded LLM calls with verification steps for structured analysis. \textsc{Agent} skills support tool-using loops and open-ended reasoning.

\paragraph{\textsc{L4} -- Storage (Evidence Store \& Knowledge Graph).}
The storage layer separates raw evidence from the graph view used for research memory. The Evidence Store is an append-only record of every artifact produced by a skill. Each artifact is stored with its lineage, so any downstream claim can be traced back to the source data that supported it. This gives the system a stable audit trail.

The Knowledge Graph turns this evidence record into a typed view of the pod's working universe. It connects tickers, memos, analysts, themes, and other research entities. It also reconstructs cross-coverage and authorship relationships from the artifact stream rather than relying on manual curation. By keeping immutable evidence separate from the derived graph, FundaPod can update the graph schema over time without rewriting the historical evidence record.

\paragraph{\textsc{L5} -- Sources.}
The source layer contains the external and internal data feeds used by the system, including SEC filings, market data, news, transcripts, alternative data, and proprietary databases. These sources enter FundaPod through ingest skills that write artifacts into the Evidence Store. Analysts do not read raw feeds directly. Instead, they work with recorded artifacts whose origin, content, and lineage are preserved. This indirection is what gives the system its traceability: every LLM-facing input is tied to a source record that can be inspected later.

\section{Core Components}
\label{sec:components}

\subsection{Persona Distillation}
\label{sec:persona}

\begin{figure}[!htbp]
\centering
\includegraphics[width=0.98\textwidth]{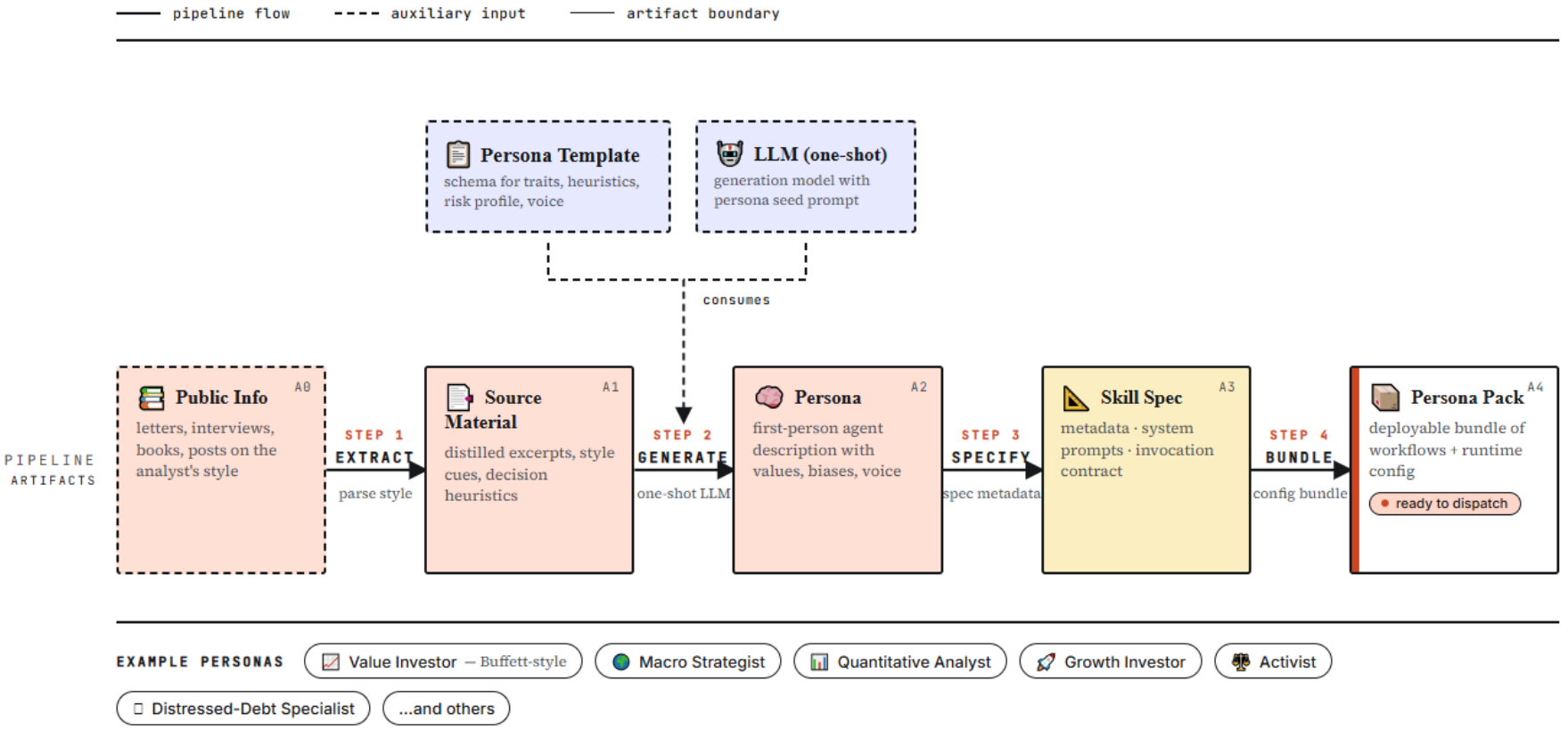}
\caption{%
\textbf{The persona distillation pipeline.}
FundaPod turns publicly available information about a known analyst or investment style, such as letters, interviews, books, and posts, into a deployable agent.
The pipeline has four steps.
Step~1 (\emph{Extract}) parses the source corpus into structured style cues and decision heuristics (\textsc{a1}).
Step~2 (\emph{Generate}) combines \textsc{a1} with a fixed Persona Template and produces a first-person Persona description (\textsc{a2}) that captures values, biases, and voice.
Step~3 (\emph{Specify}) compiles the persona into a Skill Spec (\textsc{a3}) containing metadata, system prompts, and an invocation contract.
Step~4 (\emph{Bundle}) packages the spec with runtime workflows and configuration into a Persona Pack (\textsc{a4}) that the dispatcher in Figure~\ref{fig:architecture} can schedule.
Example personas distilled by this pipeline are shown along the bottom.}
\label{fig:persona_distillation}
\end{figure}

A central design decision in FundaPod is that each agent represents a distinct investment philosophy, not just a generic role label. In real investment teams, analysts often approach the same company from different angles. A value investor may focus on business quality, competitive advantage, and price discipline. A macro strategist may focus on economic cycles, rates, and cross-asset conditions. A quantitative analyst may look for systematic patterns in financial and market data. FundaPod uses \emph{persona distillation} to make these analytical differences explicit and reusable.

Persona distillation is a four-step pipeline, shown in Figure~\ref{fig:persona_distillation}, that converts public material about an investment style into a deployable agent. The pipeline is organized as a sequence of artifacts, from \textsc{a0} to \textsc{a4}, rather than as one large prompt. This design makes each stage inspectable. A user can review the extracted source material, edit the generated persona, audit the skill specification, or replace the final configuration before the agent is added to the pod.

\paragraph{Step~1 -- Extract (\textsc{a0}$\to$\textsc{a1}).}
The process begins with a source corpus about the target investment style. This corpus may include shareholder letters, interviews, books, blog posts, or other public material. FundaPod parses these sources into structured material, including representative excerpts, style cues, and decision heuristics. Each item is indexed by the type of judgment it supports, such as business quality, valuation discipline, risk assessment, or macro sensitivity. Separating extraction from generation keeps the parsing step distinct from the LLM call. If the source corpus is corrected or expanded, the system can regenerate the persona from cleaner inputs without returning to raw text.

\paragraph{Step~2 -- Generate (\textsc{a1}$\to$\textsc{a2}).}
The structured source material is then passed to a fixed \emph{Persona Template}. The template defines the fields that every persona must contain, including traits, investment heuristics, risk profile, preferred evidence, and communication style. A single LLM call fills this template and produces a first-person \emph{Persona} description. This step captures the persona's values, biases, and tone in a format that is readable to users and usable by the system.
The fixed template is important. Without it, the LLM could decide on its own what aspects of an investor or investment style to emphasize. By fixing the fields, FundaPod makes personas comparable. For example, a Buffett-style persona and a macro-strategist persona can be compared along the same dimensions, even though they reason about companies in very different ways.

\paragraph{Step~3 -- Specify (\textsc{a2}$\to$\textsc{a3}).}
The Persona is then compiled into a \emph{Skill Spec}. The Skill Spec contains the structured metadata that the system needs to run the persona as an agent. This includes the persona's capabilities, execution phase, runner type, resource limits, system prompts, and invocation contract. The contract declares what the persona needs as input and what it produces as output.
This step separates the human-readable investment rationale from the machine-readable execution interface. The Persona explains why the agent reasons in a certain way. The Skill Spec tells the Skill Registry (\S\ref{sec:registry}) how the agent can be invoked inside a research workflow.

\paragraph{Step~4 -- Bundle (\textsc{a3}$\to$\textsc{a4}).}
Finally, the Skill Spec is packaged with default workflows, coverage sectors, and runtime configuration into a \emph{Persona Pack}. The Persona Pack is the deployable unit that the dispatcher in Figure~\ref{fig:architecture} can schedule directly. In practice, adding a new agent to the pod means adding a new Pack to the registry, rather than editing planner code.
This packaging step makes personas portable. A team can maintain a library of analyst styles and activate different combinations depending on the research question. For example, a coverage initiation may use a value investor, a growth investor, and a quantitative analyst. A macro-sensitive sector review may instead combine a macro strategist, a credit analyst, and a distressed-debt specialist.

The same pipeline produces the example roster shown at the bottom of Figure~\ref{fig:persona_distillation}: value investor, macro strategist, quantitative analyst, growth investor, activist, distressed-debt specialist, and others. Each persona carries its own analytical framework. A value-investing persona may apply a filter based on business quality, competitive moat, management integrity, and price attractiveness before committing to deeper analysis. A macro-oriented persona may instead focus on cycle position, interest-rate regimes, inflation exposure, and cross-asset correlations. By turning these styles into inspectable and deployable agents, FundaPod allows the pod to generate diverse research views without losing control over how those views are produced.

\textbf{Design rationale.}
The one-shot generation step is designed for fast research iteration rather than perfect replication of a specific investor. A hand-curated persona specification can require hours of expert review. A distilled persona can be produced in minutes and then checked by the user. This trade-off is appropriate for a research workbench, where the portfolio manager or analyst still reviews the memo and can refine the persona when the output does not match the intended style.

The artifact chain from \textsc{a1} to \textsc{a4} makes this refinement practical. Users do not need to restart from the raw source corpus. They can edit the extracted heuristics, adjust the generated Persona, revise the Skill Spec, or change the final Persona Pack. This keeps persona creation fast while preserving a clear path for human oversight and improvement.

\textbf{Skills adapter and open-source persona packs.}
Persona distillation is not the only way to add an agent to the pod. FundaPod also includes a \emph{skills adapter} for externally authored persona packs. The adapter maps these packs into the same \texttt{needs}/\texttt{produces}/phase/runner contract used by the Skill Registry. This allows personas created outside FundaPod to be scheduled by the dispatcher like native skills.

This matters because an open-source ecosystem of distilled investor skills is beginning to emerge. One example is the Buffett skill at \url{https://github.com/agi-now/buffett-skills}, a public pack that encodes Warren Buffett's investment framework, including moat analysis, management quality, valuation discipline, and capital allocation. The adapter loads such a pack, validates its declared inputs and outputs against the FundaPod schema, and registers it as a Compose-phase agent skill.

The result is an external contribution path for the persona axis (\S\ref{sec:discussion}). As practitioners publish new distilled-investor skills, FundaPod can absorb them without re-running the full persona distillation pipeline. Appendix~\ref{app:buffett_skill} describes the Buffett pack in more detail, and Appendix~\ref{app:examples} compares the same NVDA pitch with and without the Buffett skill loaded.

\subsection{Declarative Skill Registry}
\label{sec:registry}

\begin{figure}[!htbp]
\centering
\includegraphics[width=0.98\textwidth]{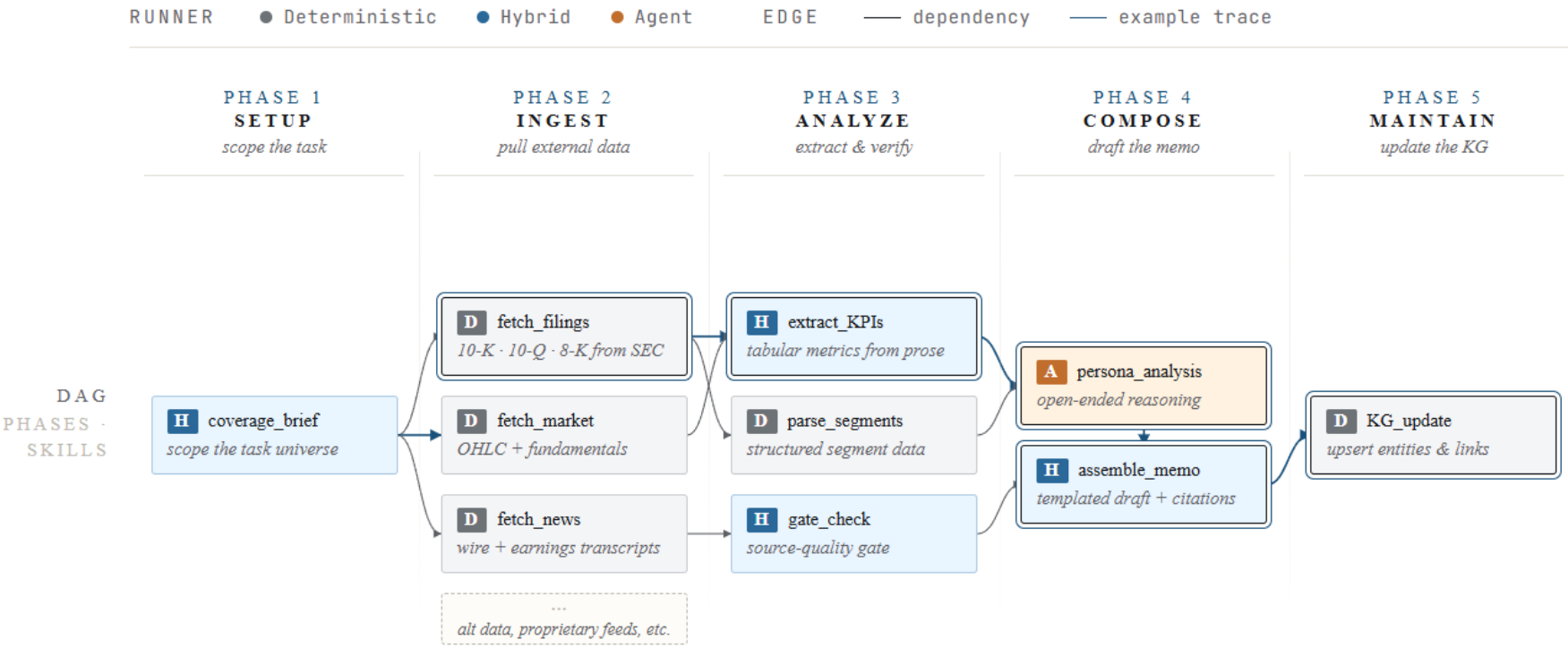}
\caption{%
\textbf{An end-to-end task as a typed skill DAG.}
Each FundaPod task is compiled into a directed acyclic graph over the Skill Registry and organized into five phases.
\emph{Setup} defines the relevant company, sector, or coverage universe.
\emph{Ingest} pulls external sources, including SEC filings, market data, news, transcripts, and additional feeds, in parallel.
\emph{Analyze} extracts KPIs, parses business segments, and applies a source-quality gate.
\emph{Compose} runs persona-conditioned reasoning and assembles a templated memo.
\emph{Maintain} writes the resulting entities and relationships back to the Knowledge Graph.
Skills are colored by runner type: \emph{Deterministic} skills for parsers and retrievers, \emph{Hybrid} skills for retrieval-grounded LLM calls with verifiers, and \emph{Agent} skills for open-ended tool-using reasoning.
The accent path traces one realized execution, while the dashed placeholder in Phase~2 shows that the Ingest phase can be extended with additional data sources.}
\label{fig:skill_dag}
\end{figure}

FundaPod uses a \emph{declarative skill registry} to make research workflows extensible. In a traditional orchestration system, developers must manually connect each data source, analysis step, and memo-writing routine. That approach becomes difficult to maintain as teams add new data vendors, sector templates, or analyst styles. FundaPod instead asks each skill to declare what it \emph{needs} as input and what it \emph{produces} as output. The planner then uses these declarations to build the research workflow automatically.
This design turns each research engagement into a typed directed acyclic graph (DAG) over registered skills. As shown in Figure~\ref{fig:skill_dag}, the graph describes the dependency structure of the task. For example, a memo-writing skill may need KPIs, filing excerpts, market data, and a persona analysis. The planner does not need to know how each input is produced in advance. It only needs to find skills that declare the corresponding outputs. This allows the same logical workflow to run across deterministic, hybrid, and agentic implementations behind a common contract.

\paragraph{Skill specification.}
Each skill is a specification with two parts.
The \emph{structured metadata} declares the skill's name, its execution phase, its runner type, the list of input categories it consumes (e.g.\ regulatory filings, market snapshots), the output categories it produces, model configuration, and resource limits.
The \emph{procedural body} is either step-by-step instructions injected into the LLM system prompt (for skills that include LLM calls) or programmatic logic executed directly (for deterministic ones).
The planner only reads the metadata; the body is opaque to the DAG-derivation logic, so changing a skill's implementation does not invalidate any graph it participates in.

\paragraph{Five execution phases.}
Skills are organised into the five sequential phases shown in Figure~\ref{fig:skill_dag}:
\emph{Setup} scopes the task universe and pulls a coverage brief;
\emph{Ingest} fetches external corpora (SEC filings, market data, news, earnings transcripts, and any number of additional feeds) in parallel;
\emph{Analyze} extracts KPIs from financial prose, parses regulatory segments, and applies a source-quality gate before further work proceeds;
\emph{Compose} runs persona-conditioned reasoning and assembles a templated memo with inline citations;
\emph{Maintain} writes the resulting entities and relations back to the Knowledge Graph (\S\ref{sec:kg}).
Phases give the DAG a natural failure boundary: a problem in Ingest aborts before any LLM tokens are spent on Compose, and a gate failure in Analyze short-circuits the rest of the engagement.

\paragraph{Three runner types behind one contract.}
Every skill, regardless of phase, is implemented as one of three runner types, and the same \texttt{needs}/\texttt{produces} contract is honoured by all three.
\emph{Deterministic} skills (parsers, retrievers, formula evaluators, e.g.\ \texttt{fetch\_filings}, \texttt{fetch\_market}, \texttt{KG\_update}) run as plain Python functions with no LLM calls, and are unit-testable in the usual way.
\emph{Hybrid} skills (retrieval-grounded LLM calls with verifiers, e.g.\ \texttt{extract\_KPIs}, \texttt{parse\_segments}, \texttt{gate\_check}, \texttt{assemble\_memo}) use an LLM under structured constraints: a fixed output schema, and a deterministic verifier that rejects malformed results. This constraint structure bounds LLM outputs to a verifiable shape, which is what makes hybrid skills reproducible despite the underlying model's stochasticity.
\emph{Agent} skills (tool-using loops, open-ended reasoning, e.g.\ \texttt{persona\_analysis}) invoke an LLM with tool access and procedural instructions when the work is genuinely open-ended.
The contract is the only thing the registry actually exposes. A downstream skill that declares it needs \texttt{kpis} does not see whether they were produced by a hybrid extractor today or a deterministic parser tomorrow.

\paragraph{Automatic DAG derivation.}
When the planner constructs a task list for an engagement, it starts from the chosen Compose skill, reads its declared input categories, looks up which Analyze and Ingest skills produce each one, and walks the dependency graph backwards until every leaf is a Setup-phase skill.
The result is the typed DAG in Figure~\ref{fig:skill_dag}, executed in topological order with idempotent retry semantics (\S\ref{sec:planner}).
Adding a new data-source skill with an appropriate output declaration makes it instantly available to any Compose skill that declares the corresponding input, without any planner code changes.

The skill registry is designed to be extensible across data sources.
FundaPod can ingest data from regulatory filing databases (e.g., SEC EDGAR), market data APIs, news aggregators, earnings transcript services, alternative data providers, and proprietary internal databases.
Each data source is wrapped in a deterministic skill with appropriate metadata, making it composable with any downstream analysis or composition skill.

\subsection{Task Planner and Dispatcher}
\label{sec:planner}

The task planner and dispatcher form the execution backbone of FundaPod.
They convert high-level engagement requests (e.g., ``write a pitch memo on ticker X'') into ordered, dependency-respecting task sequences.

\textbf{Planning.}
The planner is template-driven. Each engagement type (pitch memo, coverage update, thematic exploration) maps to a plan template that defines the required phases and skills.
Given a template and target, the planner scans the data-source registry, emits ingest tasks for each data dependency, adds analysis and composition tasks, and outputs a task list with explicit dependencies.

\textbf{Dispatching.}
The dispatcher topologically sorts tasks and executes them in phase order.
Each task transitions through a lifecycle: pending $\to$ in-progress $\to$ done $|$ error $|$ skipped.
Completed tasks are skipped on re-execution, providing idempotent retry semantics.
The dispatcher supports real-time progress updates via Server-Sent Events to the user interface.

\textbf{Conditional routing.}
Some skills implement conditional logic based on intermediate results.
For instance, a value-investing persona may apply a quick screening filter before committing to full analysis: if the target fails the initial screen, the engagement terminates early with a brief assessment, saving compute and analyst time.
This routing is driven by task parameters and skill-internal logic, not by the planner.

\subsection{Grounded Evidence Model}
\label{sec:evidence}

FundaPod adopts a \emph{grounded evidence model}: every intermediate and final artifact produced during a research engagement is persisted with full provenance metadata.
When an agent skill generates analytical text, it references source artifacts by stable identifiers, enabling a reader (human or machine) to trace any claim back to the specific data (a regulatory filing section, a financial metric, a news article) that supports it.
This design contrasts with vector-database retrieval-augmented generation (RAG) systems, where retrieved context is opaque: the user cannot inspect which document chunks were used or how similarity scores determined relevance.
In FundaPod, the evidence chain is explicit and auditable.

The design rationale behind the grounded evidence model is that investment research must be defensible after the memo is written. \textbf{Auditability} means that a portfolio manager, risk officer, or investment committee can trace a claim back to the filing, data table, transcript, news article, or internal note that supported it. This makes source verification a first-order system requirement rather than an afterthought. The same evidence structure also supports \textbf{version control}. Because the system stores the full evidence history, prior research can be versioned and reconstructed when assumptions, data sources, or models change. It also improves \textbf{tool compatibility}: analysts can inspect, search, and share the underlying artifacts using standard tools, rather than relying only on opaque retrieval results from an indexed store.

The trade-off is that cross-engagement queries require the Knowledge Graph (Section~\ref{sec:kg}) rather than a single indexed store. FundaPod accepts this trade-off because it separates two different needs: the Evidence Store preserves source-level auditability, while the Knowledge Graph supports higher-level queries across companies, analysts, memos, and themes.

\subsection{Knowledge Graph Second Brain}
\label{sec:kg}

\begin{figure}[!htbp]
\centering
\includegraphics[width=0.98\textwidth]{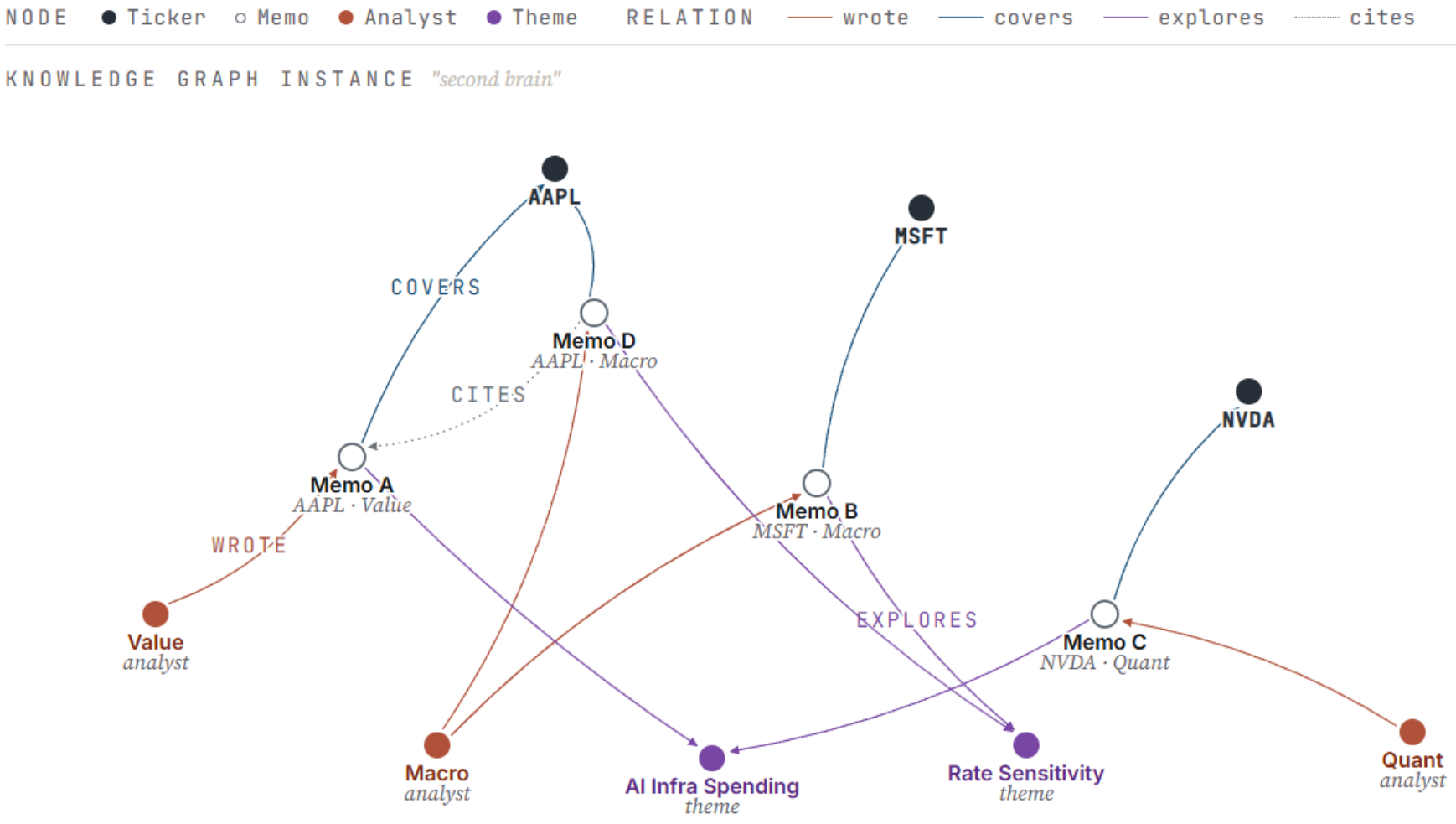}
\caption{%
\textbf{The knowledge-graph ``second brain.''}
FundaPod maintains a typed property graph over the pod's working universe.
Four node types (\emph{Ticker}, \emph{Memo}, \emph{Analyst}, \emph{Theme}) are connected by four directed relations: an Analyst \emph{wrote} a Memo; a Memo \emph{covers} one or more Tickers and \emph{explores} one or more Themes; and Memos \emph{cite} other Memos when they build on prior work.
The illustrated subgraph contains three tickers, four memos, three personas, and two themes.
Memo~D, written by the Macro persona about AAPL with a rate-sensitivity angle, \emph{cites} the earlier Memo~A on the same ticker, demonstrating how the graph accumulates cross-memo provenance over the life of the deployment.}
\label{fig:knowledge_graph}
\end{figure}

As the pod accumulates research across multiple analysts, tickers, and themes, individual artifact-level navigation becomes insufficient.
FundaPod addresses this with a \emph{knowledge graph} that projects the pod's collective research state into a typed property graph (Figure~\ref{fig:knowledge_graph}).
We use the phrase \emph{second brain} in the sense Karpathy uses it in his LLM-Wiki note~\citep{karpathy2026llmwiki}: a persistent, compounding artifact that an LLM builds and maintains over time rather than re-deriving on every query, so that the agent's prior reasoning becomes addressable knowledge rather than a transient context window.
The illustrated subgraph shows the pattern the graph is designed to capture. When the Macro persona writes Memo~D on AAPL with a rate-sensitivity angle, the graph records that Memo~D \emph{covers} AAPL, \emph{explores} the rate-sensitivity theme, and \emph{cites} the earlier Value-persona Memo~A on the same ticker. The new memo's reasoning chain back to prior coverage is therefore queryable rather than implicit.

\textbf{Node types.}
The graph contains four node types: \emph{Ticker} (equity symbols under coverage, e.g.\ \textsc{aapl}, \textsc{msft}, \textsc{nvda} in the illustrated subgraph), \emph{Memo} (individual analyst research outputs), \emph{Analyst} (persona agents such as Value, Macro, and Quant), and \emph{Theme} (cross-cutting investment ideas such as ``AI Infra Spending'' or ``Rate Sensitivity'').
Memo nodes carry the metadata of the engagement that produced them (covered ticker, authoring persona, timestamp), so a memo's identity is not just a string label but a typed reference into the Evidence Store.

\textbf{Edge extraction.}
Edges are derived from the evidence store:
\emph{wrote} edges connect analysts to their memos;
\emph{covers} edges connect memos to tickers;
\emph{explores} edges connect memos to themes;
\emph{cites} edges connect memos that reference each other's artifacts.
All edges are extracted automatically from structured metadata and cross-reference patterns in generated text.

\textbf{Stateless reconstruction.}
The graph is reconstructed on each query from the current state of the evidence store.
This stateless design ensures consistency without migration or synchronization logic.
For the target scale of a single portfolio manager's pod (tens to low hundreds of engagements), reconstruction stays within an acceptable latency budget for interactive use, without requiring incremental-update infrastructure.

\textbf{Use cases.}
In terms of use cases, the Knowledge Graph serves as a ``second brain'' for the portfolio manager. It turns the pod's research output into a shared map of companies, analysts, memos, themes, and supporting evidence. This gives the PM a way to move from a high-level view of coverage to the underlying work that supports each conclusion.

One use case is \textbf{gap detection}. If a ticker has been reviewed by only one persona, the graph makes that limited coverage visible. The PM can then assign another persona to review the same company from a different angle, such as adding a macro review to a value-oriented memo or a quantitative screen to a growth thesis.

A second use case is \textbf{thematic synthesis}. Theme nodes aggregate memos across agents and tickers. This allows the PM to see which parts of the existing research base relate to a given investment thesis, such as AI infrastructure, consumer weakness, margin pressure, or refinancing risk. Instead of searching memo by memo, the PM can start from the theme and inspect the relevant companies, analysts, and evidence.

A third use case is \textbf{provenance navigation}. Each graph node connects back to the artifacts that produced it. Clicking a node lets the PM drill down from the portfolio-level view to the underlying memo, data table, filing excerpt, transcript, or other source artifact. This makes the graph not only a summary layer, but also an entry point into the evidence trail.

This design relates to recent work on graph-based agent memory~\citep{yang2026graphbasedagentmemorytaxonomy, magma2026, wu2026gamhierarchicalgraphbasedagentic}, but it serves a different purpose. Rather than supporting one agent's episodic recall, FundaPod uses the graph to synthesize \emph{multi-agent} research outputs into a shared view for the master agent and the PM. The personas do not read this shared graph during their own reasoning. This separation preserves the pod's independence property. In other words, the Knowledge Graph is a vantage point for the PM and the master agent, not a back-channel between agents. It helps the system compare views after they are formed, without allowing one persona's reasoning to shape another's analysis too early.
\section{Case Study: Research Engagement}
\label{sec:case_study}

\begin{figure}[!htbp]
\centering
\includegraphics[width=0.98\textwidth]{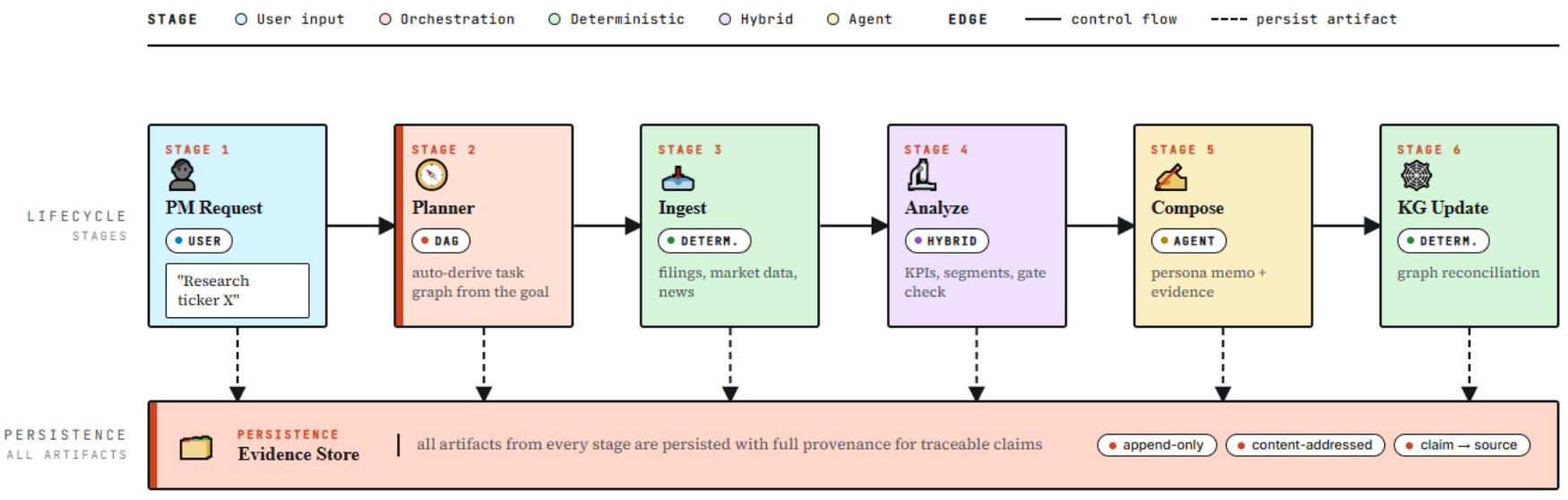}
\caption{%
\textbf{The research engagement lifecycle.}
A typical FundaPod engagement moves through six stages.
A portfolio manager begins with a natural-language request, such as \emph{``Research ticker X''}.
The Planner turns that request into a typed task graph over the Skill Registry.
The system then ingests filings, market data, and news; analyzes the material by extracting KPIs, parsing business segments, and checking source coverage; and composes a memo through persona-conditioned reasoning.
After the memo is delivered, the Knowledge Graph is updated so the new work becomes part of the pod's research memory.
The Evidence Store supports the full lifecycle. Every intermediate artifact is stored with provenance and a stable identifier, so each claim in the final memo can be traced back to its source data.}
\label{fig:engagement_flow}
\end{figure}

Figure~\ref{fig:engagement_flow} illustrates a complete FundaPod research engagement. The example begins with a portfolio manager asking the system to research a target equity using a selected persona, such as a value-investing persona covering a large-cap technology company. From that request, FundaPod creates an engagement context and starts a research workflow that moves from planning to ingestion, analysis, memo composition, and knowledge-graph integration.

The first stage is \textbf{Engagement Creation}. The portfolio manager's request is treated as a research goal rather than as a single prompt. The system records the target company, the selected persona, and the requested research objective. It then passes this context to the planner, which selects the appropriate research template for the engagement.

The second stage is \textbf{Task Planning}. The planner reads the metadata for the memo-composition skill and identifies the inputs required to produce the final research output. These inputs include a coverage brief, regulatory filings, market data, news, parsed financial segments, and extracted KPIs. The planner then searches the skill registry for producer skills that can supply each input. The result is a task graph that spans the main execution phases: setup, ingest, analyze, compose, and maintain.

Once the task graph is created, the engagement moves to \textbf{Data Ingestion}. The dispatcher runs the ingest tasks in dependency order. It retrieves regulatory filings, collects market data snapshots, and aggregates relevant news. These steps are deterministic and do not require LLM calls. Each output is written to the Evidence Store with provenance metadata, so later stages can identify where the data came from and how it entered the workflow.

The next stage is \textbf{Analysis}. Here the system converts raw source material into research-ready inputs. Extraction skills parse financial statements to identify key operating metrics, such as revenue trends, margin profiles, and growth rates. Segment-parsing skills extract management discussion, business overview, and risk disclosures from regulatory filings. A coverage gate then reviews the extracted material for completeness before the workflow proceeds to memo composition. This stage combines structured processing with LLM-assisted interpretation, but it remains grounded in recorded evidence.

The engagement then moves to \textbf{Composition}. The persona skill receives the full engagement context, including the coverage brief, extracted KPIs, parsed filing segments, market data, and news. The selected persona applies its analytical framework to the company. For example, a value-investing persona may focus on business quality, competitive positioning, management, capital allocation, and valuation. The system then assembles the resulting sections into a final research memo, with source references that link each material claim to its supporting artifact.

The final stage is \textbf{Knowledge Graph Integration}. After the memo is completed, the Knowledge Graph is rebuilt or updated from the Evidence Store. The new memo becomes a node connected to the target ticker, the authoring analyst, the source artifacts, and any themes referenced in the analysis. If another persona has previously researched the same company, both memos connect to the same ticker node. This allows a portfolio manager to compare analytical perspectives, review prior coverage, and see how the pod's view of the company has evolved over time.

\section{Discussion}
\label{sec:discussion}

\begin{table*}[t]
\centering
\caption{Comparison of Compass with existing financial AI agent systems and multi-agent frameworks. \cmark\ = fully supported, \pmark\ = partially supported, \xmark\ = not supported. All claims are grounded in the cited papers.}
\label{tab:comparison}
\resizebox{\textwidth}{!}{%
\begin{tabular}{lccccccc}
\toprule
\textbf{System} & \textbf{Multi-Persona} & \textbf{Evidence} & \textbf{Knowledge} & \textbf{Skill} & \textbf{Workflow} & \textbf{Target} & \textbf{Domain} \\
 & \textbf{Support} & \textbf{Grounding} & \textbf{Graph} & \textbf{Composability} & \textbf{Automation} & \textbf{Use Case} & \\
\midrule
\multicolumn{8}{l}{\textit{Financial LLM Models \& Agents (2023--2024)}} \\
BloombergGPT \cite{wu2023bloomberggpt} & \xmark\ Single model & \xmark\ No traceability & \xmark\ None & \xmark\ Monolithic & \xmark\ Manual & NLP tasks & Finance \\
FinGPT \cite{yang2025fingptopensourcefinanciallarge} & \xmark\ Single model & \xmark\ No traceability & \xmark\ None & \xmark\ Pipeline-based & \pmark\ Data pipeline & Trading / NLP & Finance \\
FinAgent \cite{zhang2024finagent} & \xmark\ Single agent & \pmark\ Memory retrieval & \xmark\ None & \pmark\ Tool integration & \pmark\ Reflection loop & Trading signals & Finance \\
FinMem \cite{yu2025finmem} & \pmark\ Character design & \pmark\ Layered memory & \xmark\ None & \xmark\ Fixed pipeline & \pmark\ Memory-driven & Trading signals & Finance \\
\midrule
\multicolumn{8}{l}{\textit{Financial Agent Systems (2025--2026)}} \\
TradingAgents \cite{xiao2025tradingagentsmultiagentsllmfinancial} & \cmark\ Role-based team & \pmark\ Shared memory & \xmark\ None & \pmark\ Tool use & \cmark\ Debate loop & Trading signals & Finance \\
Expert Teams \cite{miyazaki2026expertteams} & \cmark\ Fine-grained roles & \pmark\ Task outputs & \xmark\ None & \pmark\ Task-specific & \cmark\ Task pipeline & Trading tasks & Finance \\
FinSaber \cite{finsaber2026} & \xmark\ Single strategy & \pmark\ Backtest logs & \xmark\ None & \pmark\ Strategy tools & \pmark\ Backtest loop & Long-run trading & Finance \\
\midrule
\multicolumn{8}{l}{\textit{General Multi-Agent Frameworks}} \\
AutoGen \cite{wu2024autogen} & \pmark\ Role assignment & \xmark\ No artifacts & \xmark\ None & \pmark\ Code execution & \cmark\ Conversation & General tasks & General \\
MetaGPT \cite{hong2023metagpt} & \cmark\ SOP roles & \pmark\ Structured outputs & \xmark\ None & \pmark\ Hardcoded SOPs & \cmark\ Assembly line & Software dev & General \\
ChatDev \cite{qian2023chatdev} & \cmark\ Company roles & \xmark\ No artifacts & \xmark\ None & \xmark\ Fixed phases & \cmark\ Chat chain & Software dev & General \\
Generative Agents \cite{park2023generativeagents} & \cmark\ 25 personas & \pmark\ Memory stream & \xmark\ None & \xmark\ No skills & \xmark\ Autonomous & Social sim. & General \\
Manus \cite{manus2025} & \xmark\ Single agent & \pmark\ Tool outputs & \xmark\ None & \cmark\ Dynamic tools & \cmark\ Autonomous & General tasks & General \\
\midrule
\multicolumn{8}{l}{\textit{Memory \& Knowledge Graph Agents}} \\
MemGPT \cite{packer2023memgpt} & \xmark\ Single agent & \pmark\ Tiered memory & \xmark\ None & \pmark\ Function calls & \xmark\ Manual & Long-context & General \\
Voyager \cite{wang2023voyager} & \xmark\ Single agent & \xmark\ No traceability & \xmark\ None & \cmark\ Skill library & \pmark\ Curriculum & Game playing & Gaming \\
AriGraph \cite{anokhin2024arigraph} & \xmark\ Single agent & \pmark\ Episodic memory & \cmark\ KG world model & \xmark\ No skills & \xmark\ Exploration & Text games & Gaming \\
MAGMA \cite{magma2026} & \xmark\ Single agent & \pmark\ Multi-graph & \cmark\ Multi-graph & \xmark\ No skills & \xmark\ Manual & General memory & General \\
GAM \cite{wu2026gamhierarchicalgraphbasedagentic} & \xmark\ Single agent & \pmark\ Hierarchical & \cmark\ Hierarchical KG & \xmark\ No skills & \xmark\ Manual & General memory & General \\
\midrule
\textbf{Compass (ours)} & \cmark\ \textbf{Distilled pods} & \cmark\ \textbf{Artifact paths} & \cmark\ \textbf{Second brain} & \cmark\ \textbf{Declarative registry} & \cmark\ \textbf{Auto DAG} & \textbf{Fundamental} & \textbf{Finance} \\
 & & & & & & \textbf{research} & \\
\bottomrule
\end{tabular}%
}
\end{table*}

\textbf{Comparison with existing systems.}
Table~\ref{tab:comparison} compares FundaPod with existing systems across seven dimensions.
Within this comparison set, FundaPod is the only system that combines all six capabilities: multi-persona agents with distilled domain expertise, artifact-level evidence grounding, knowledge graph memory, declarative skill composition, automatic workflow orchestration, and targeting of fundamental research rather than trading. The coding reflects our reading of the cited works and is intended to position FundaPod within the literature, not to settle every borderline cell.
Among recent financial agent systems, TradingAgents~\citep{xiao2025tradingagentsmultiagentsllmfinancial} is closest in its use of specialized analyst roles, but it targets short-term trading decisions rather than institutional research workflows, lacks a knowledge graph for cross-engagement synthesis, and, most relevant to our architectural argument, employs a debate-converging protocol that we deliberately reject in favor of an independence-preserving design (see \emph{Independence vs.\ inter-agent debate} below).

\textbf{Design trade-offs.}
The principles in \S\ref{sec:design_principles} entail three engineering trade-offs.
(1) \emph{Grounded evidence vs.\ vector retrieval}: The grounded evidence model (DP3) gives up some flexibility from similarity-based retrieval in exchange for full traceability. For institutional research, where every claim should be verifiable, this is the right trade-off. The Knowledge Graph then adds a higher-level query interface on top of the same evidence artifacts.
(2) \emph{One-shot distillation vs.\ fine-tuning}: Persona distillation through a single LLM call is fast and easy to extend, but it is also approximate. Fine-tuning a model on an investor's writings and interviews would likely improve persona fidelity, but it would add substantial data and engineering cost. The one-shot approach is therefore a pragmatic default, while users can manually refine generated specifications when higher accuracy is needed.
(3) \emph{Stateless graph rebuild vs.\ incremental updates}: Rebuilding the Knowledge Graph on every query avoids synchronization complexity, but it creates latency that grows with the size of the evidence store. At the target scale of a single PM's pod, this cost is acceptable. A firm-wide deployment would require incremental updates.

\paragraph{Independence vs.\ inter-agent debate (DP2 in operation).}
The most consequential architectural commitment in FundaPod is the one made explicit by DP2 (\S\ref{sec:design_principles}): persona-distilled agents reason in isolation and never observe one another's intermediate reasoning. A natural alternative would be to let personas exchange context through shared memory, or to drive them through a debate protocol until they converge on a joint view. We reject this alternative for three reasons. First, the analytical value of a persona-distilled agent is the \emph{consistent application of one investment lens}; once agents observe one another too early, that consistency degrades into a hybrid framing that no single persona endorses. Second, informational-cascade theory~\citep{bikhchandani1992cascades} predicts that once agents observe one another's actions, private signals stop being reflected in the aggregate, so the pod appears diverse while moving toward premature consensus. Third, controlled studies of LLM multi-agent debate find that much of the apparent benefit of debate protocols comes from independent-output ensembling rather than from debate itself, and that additional debate rounds can harden errors among similar agents~\citep{smit2024mad}. Among recent financial agent systems, this is also the central architectural contrast: TradingAgents~\citep{xiao2025tradingagentsmultiagentsllmfinancial} drives specialized analyst roles through a debate loop to converge on a trading signal, whereas FundaPod is designed to \emph{produce} disagreement and route it to the PM for adjudication. The output of FundaPod is therefore not a consensus view but a set of provenance-grounded, mutually inspectable persona views that the human expert combines into the final research conclusion.

\textbf{Extensibility.}
The declarative-contract design makes FundaPod straightforward to extend along four independent axes, each modifiable without touching the others:
(1) \emph{New data sources}: ingesting an additional feed (alternative data such as satellite imagery, web-traffic panels, or credit-card transactions, but also new fundamentals vendors, regulatory archives, or proprietary internal databases) requires a single new deterministic skill with the appropriate \texttt{produces} declaration. Every downstream consumer that already declares the matching \texttt{needs} category picks it up automatically.
(2) \emph{New skills}: any new capability (a fresh KPI extractor, a domain-specific verifier, a different memo template, a graph-update procedure) is added by dropping a Skill Spec into the registry. The planner discovers it through its metadata rather than through hand-written orchestration, so adding a skill never requires editing the planner or the dispatcher.
(3) \emph{New personas}: running the persona distillation pipeline (\S\ref{sec:persona}) on a new corpus of letters, interviews, or strategy descriptions produces a deployable Persona Pack; shipping a new agent to the pod means dropping that Pack into the registry, not editing any code.
(4) \emph{New engagement workflows}: defining a new plan template that names the required Compose skill and execution phases creates a new workflow. Pitch memos, coverage updates, thematic deep-dives, monitoring briefs, and post-mortems are all instances of the same DAG-derivation machinery applied to different templates.
Because each axis is mediated by the same needs/produces/phase/runner contract, extensions along one axis do not invalidate work along another: a new data source becomes immediately useful to every existing persona, a new persona inherits every existing data source, and a new workflow can mix and match both.

\textbf{Limitations.}
The current system has four main limitations. (1) \emph{No quantitative evaluation}: This paper presents a systems description rather than a controlled empirical evaluation. We do not yet measure persona differentiation quality, memo accuracy, or investment signal value in a formal benchmark. We plan to address this in follow-up work through blind evaluations by professional analysts. (2) \emph{LLM dependency}: Agent skills depend on the capabilities of the underlying LLM and remain subject to known limitations, including context-window constraints, occasional hallucination, and inference cost. FundaPod mitigates these risks through the grounded evidence model and by separating deterministic skills from agent skills. (3) \emph{Single-PM scale}: The current design targets a single portfolio manager's pod. Firm-wide deployment would require additional infrastructure for access controls, concurrent engagement management, and incremental knowledge graph updates. (4) \emph{Persona fidelity}: The one-shot distillation approach captures broad investment philosophy, but it may miss more subtle aspects of reasoning style. Formal evaluation of persona faithfulness is therefore left to future work.

\section{Conclusion and Future Work}
\label{sec:conclusion}

We have presented \textbf{FundaPod}, a multi-persona agent pod architecture for AI-driven fundamental investment research. The system combines persona-distilled agents, a declarative skill registry, a grounded evidence model, and a knowledge graph ``second brain.'' Together, these components address several requirements of institutional research: domain expertise, repeatable workflows, evidentiary rigor, and synthesis across companies, sectors, and themes.

FundaPod is designed to support the research process rather than replace it with a single model prediction. Its goal is to help investment teams collect evidence, apply distinct analytical lenses, produce traceable memos, and preserve coverage knowledge over time. By separating deterministic skills from LLM-driven agent skills, the architecture keeps data processing reproducible while using language models for judgment-heavy tasks such as memo writing, thematic exploration, and thesis comparison.

Future work will focus on evaluation, deployment, and richer market integration. First, we plan to conduct blind evaluations of memo quality, factual grounding, and persona differentiation with professional equity analysts. Second, we will develop quantitative metrics for knowledge graph coverage, including coverage completeness, entity linkage quality, and gap detection accuracy. Third, we will extend the system for multi-PM and firm-wide deployment, with access controls, incremental graph updates, and support for team-specific research workflows. Fourth, we will evaluate whether FundaPod-generated research provides investment signal value relative to human analyst benchmarks. Finally, we will explore integration with dynamic financial knowledge graphs~\citep{findkg2024}, so that real-time market events can be linked to the pod's existing research memory.

FundaPod also opens several directions for future research. Subsequent work can study how multi-persona research systems should be evaluated, how evidence-grounded memos affect analyst trust, and how knowledge graph memory changes collaboration across investment teams. More broadly, our results suggest that AI systems for finance should be evaluated not only by prediction accuracy or trading performance, but also by their ability to support transparent, reusable, and auditable research workflows.

\bibliographystyle{plainnat}
\bibliography{references}

\newpage
\appendix

\section{Application Screenshots}
\label{app:screenshots}

This appendix shows the running FundaPod user interface across its principal views.
The screenshots correspond to the architectural components described in \S\ref{sec:architecture}--\S\ref{sec:components}: the dashboard exposes the Analyst Pod and recent memos; the Talent Pool, Skills library, Workflows library, and Data library expose the four extension axes (personas, skills, workflows, data sources; see \S\ref{sec:discussion}); the Knowledge Graph view renders the ``second brain'' described in \S\ref{sec:kg}.

\begin{figure}[!htbp]
\centering
\includegraphics[width=\textwidth]{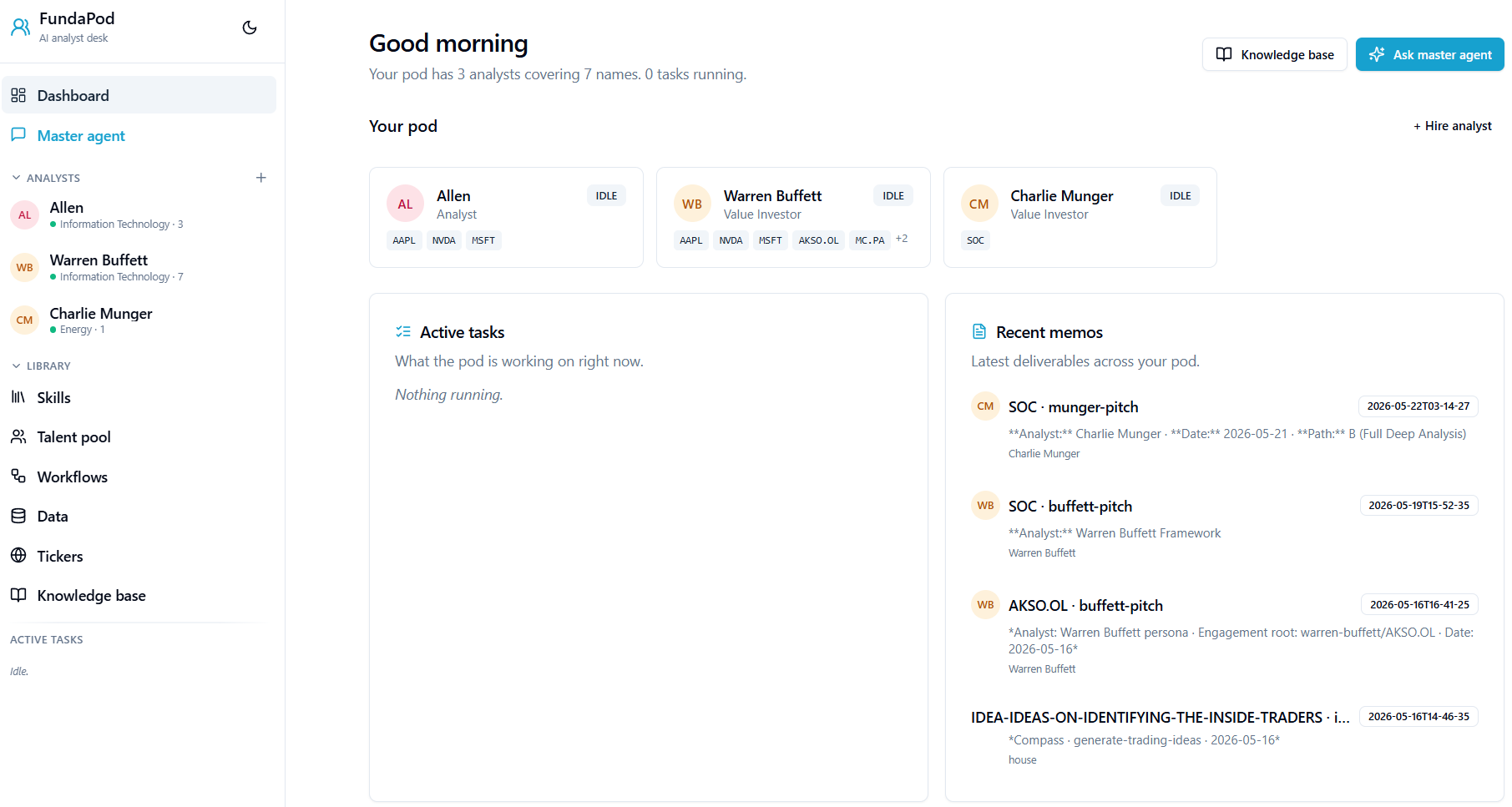}
\caption{%
\textbf{Dashboard view.}
The pod overview surfaces the persona-distilled agents currently in the pod (here Allen, Warren Buffett, and Charlie Munger; the UI labels them as ``hired analysts''), their coverage tickers, currently running tasks, and recently produced memos.
The left navigation gives the portfolio manager access to the Master Agent (the only component that sees across agents) and to the four library views (Skills, Talent pool, Workflows, Data) plus per-ticker and knowledge-base entry points.}
\label{fig:ui_dashboard}
\end{figure}

\begin{figure}[!htbp]
\centering
\includegraphics[width=\textwidth]{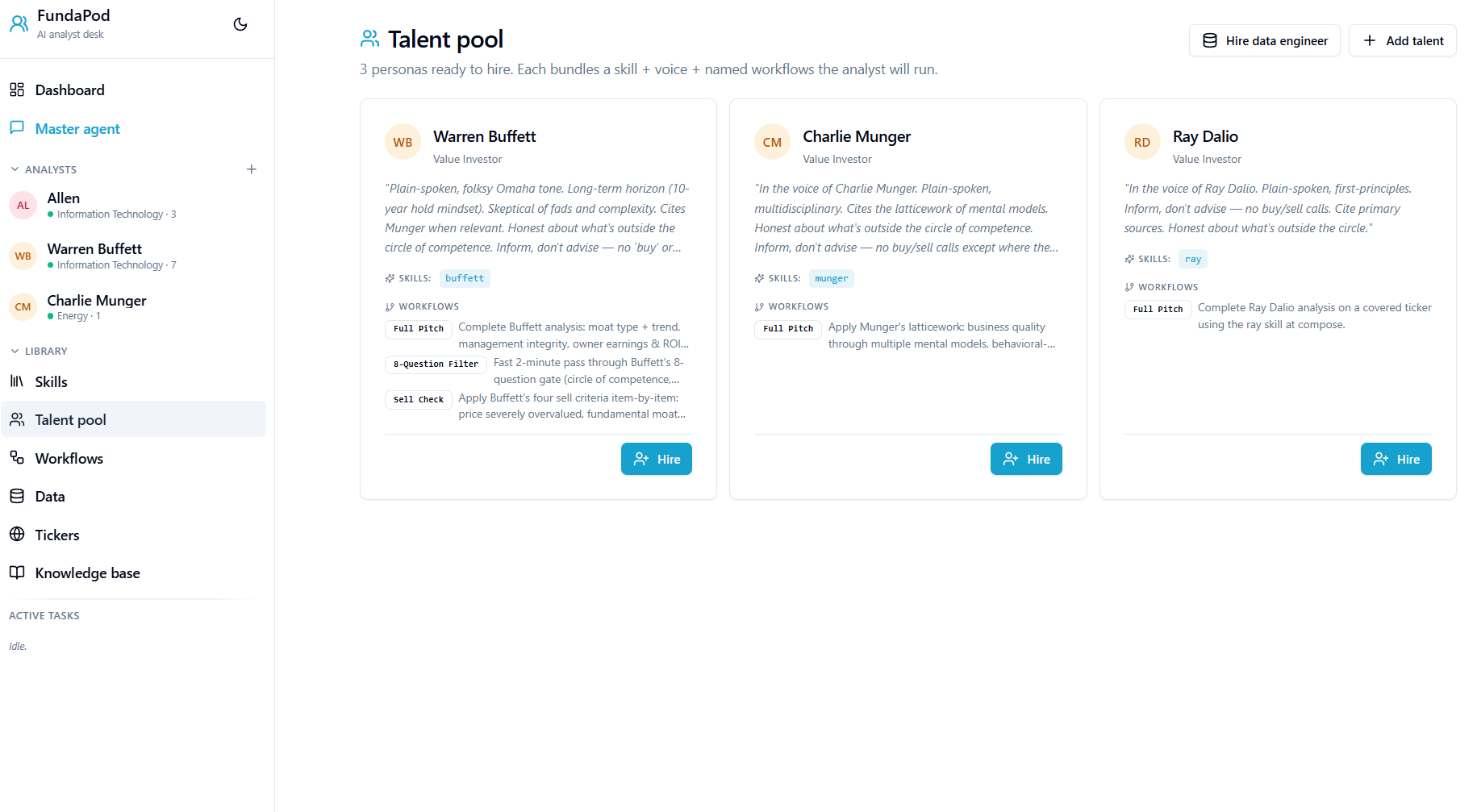}
\caption{%
\textbf{Talent pool.}
Three persona-distilled agents (Warren Buffett, Charlie Munger, Ray Dalio) sit ready to hire.
Each card shows the persona's voice in a short quote, the underlying skill, and the named workflows it can run (Full Pitch, 8-Question Filter, Sell Check).
``Hire data engineer'' and ``Add talent'' are the entry points for the persona-extension axis (\S\ref{sec:persona}).}
\label{fig:ui_talent_pool}
\end{figure}

\begin{figure}[!htbp]
\centering
\includegraphics[width=\textwidth]{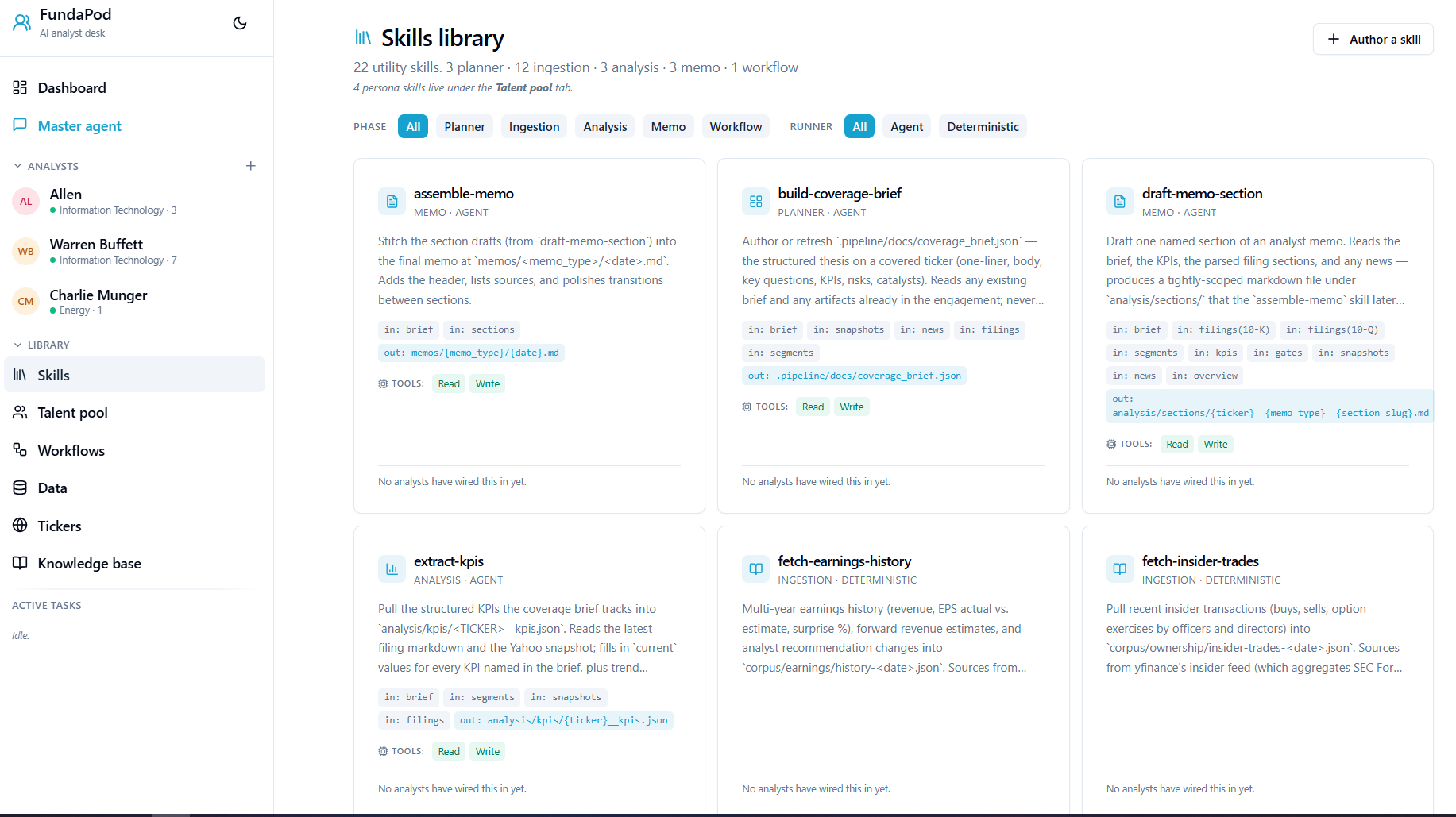}
\caption{%
\textbf{Skills library.}
Twenty-two skills are visible across the five execution phases (Planner, Ingestion, Analyze, Memo, Workflow) and three runner types (Agent, Deterministic, and the hybrid combinations).
Each card declares the skill's contract (producer/consumer categories and runner type), which is what the planner reads when deriving a DAG (Figure~\ref{fig:skill_dag}).}
\label{fig:ui_skills}
\end{figure}

\begin{figure}[!htbp]
\centering
\includegraphics[width=\textwidth]{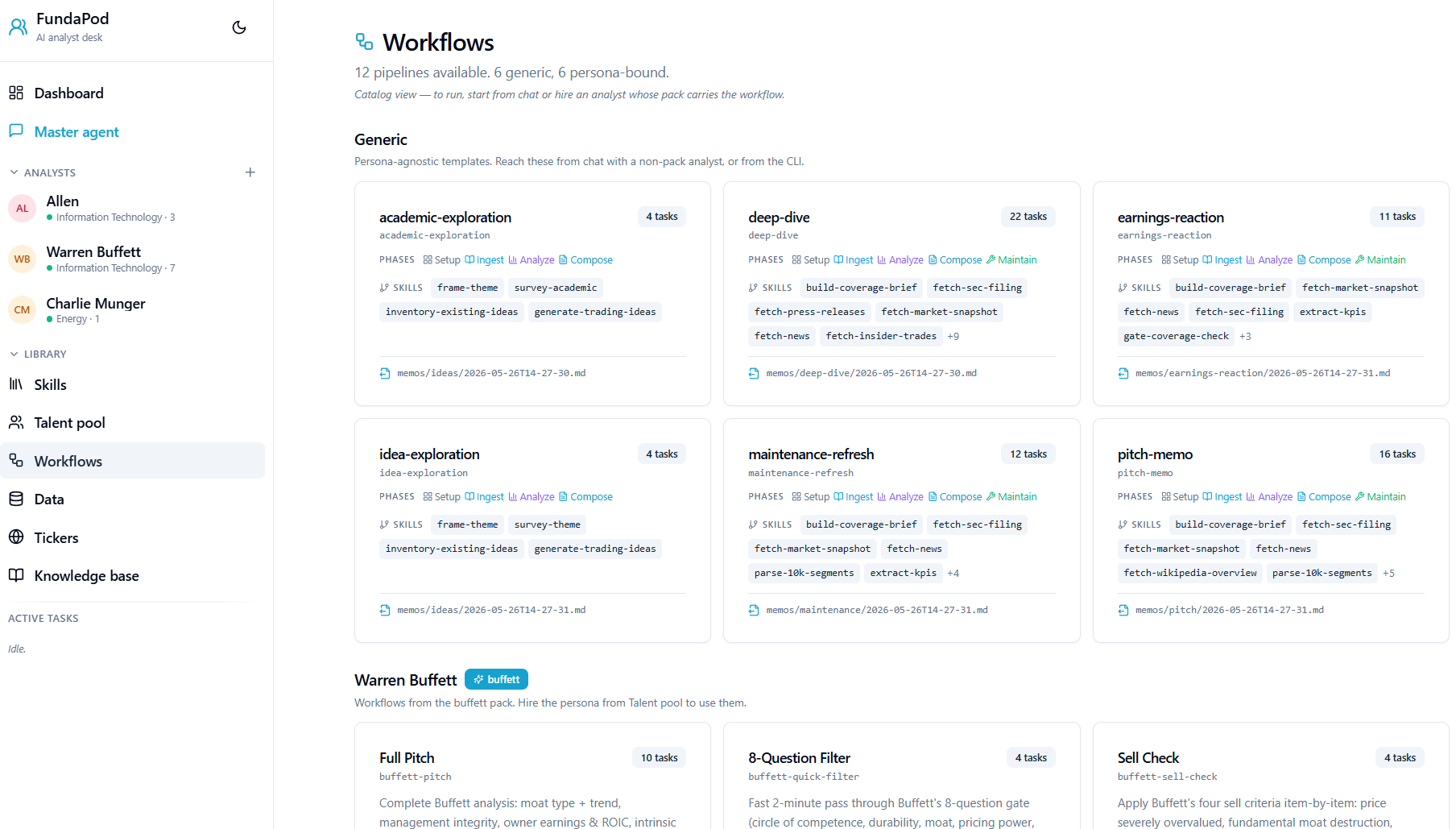}
\caption{%
\textbf{Workflows library.}
Eleven pipelines, six persona-agnostic templates (academic-exploration, deep-dive, earnings-reaction, idea-exploration, maintenance-refresh, pitch-memo) and five persona-bound workflows (e.g.\ Full Pitch, 8-Question Filter, Sell Check under the Warren Buffett pack), each list their phases and constituent skills.
Adding a new workflow is a matter of dropping a template into this view; the dispatcher then schedules it through the same DAG-derivation machinery (\S\ref{sec:planner}).}
\label{fig:ui_workflows}
\end{figure}

\begin{figure}[!htbp]
\centering
\includegraphics[width=\textwidth]{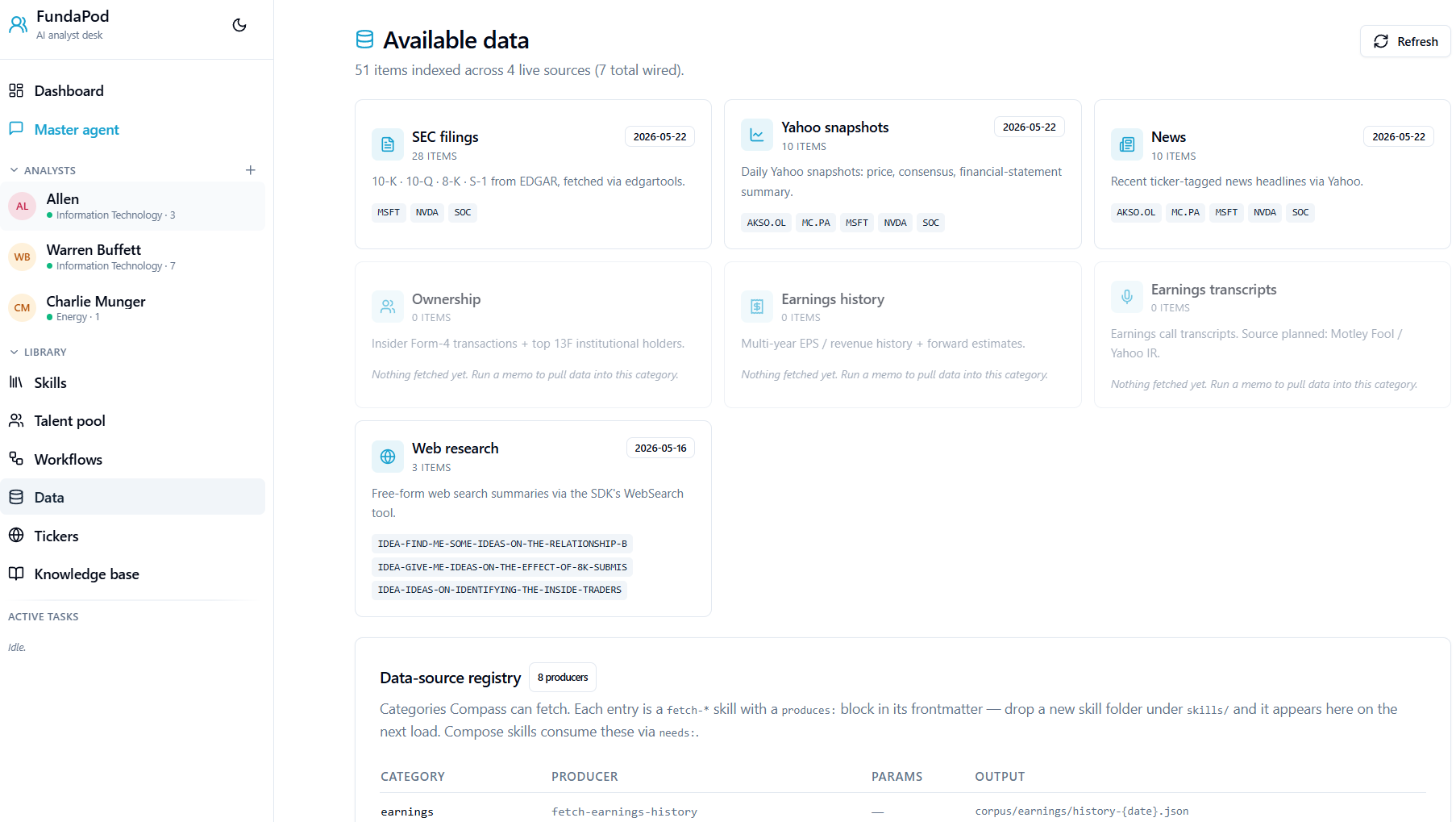}
\caption{%
\textbf{Available data.}
The data library lists current items per source: SEC filings, Yahoo snapshots, news, web research, plus three further categories (ownership, earnings history, earnings transcripts) that are wired but not yet populated.
The lower ``Data-source registry'' panel renders the same producer registry the planner consults: each entry is a \texttt{fetch-*} skill with a \texttt{produces:} declaration whose category becomes immediately consumable by any downstream skill that declares the corresponding \texttt{needs:} (\S\ref{sec:registry}).}
\label{fig:ui_data}
\end{figure}

\begin{figure}[!htbp]
\centering
\includegraphics[width=\textwidth]{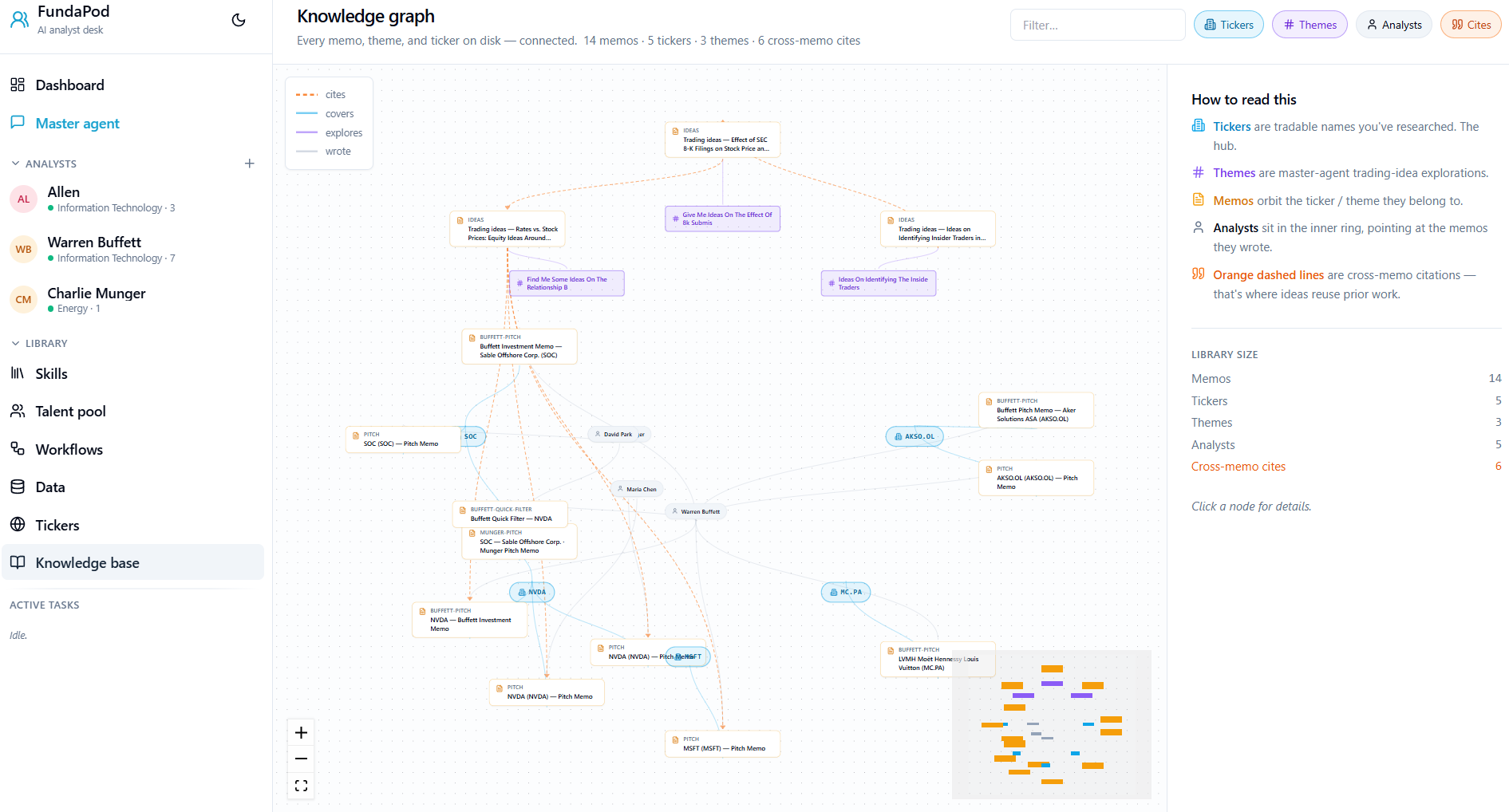}
\caption{%
\textbf{Knowledge-graph ``second brain.''}
The interactive graph view materialises the property graph described in \S\ref{sec:kg}: tickers, memos, personas, and themes connected by \emph{covers}, \emph{wrote}, \emph{explores}, and \emph{cites} edges.
The portfolio manager can pan and zoom across the pod's accumulated coverage; the right-hand panel summarises library size (memos, tickers, themes) and offers \emph{Today}, \emph{Memos}, and \emph{Analyze} entry points.
As noted in \S\ref{sec:kg}, this view is consumed by the master agent and the PM, not by the personas, preserving the pod's independence property.}
\label{fig:ui_second_brain}
\end{figure}

\section{Onboarding external persona skills}
\label{app:buffett_skill}

The persona-extension axis (\S\ref{sec:discussion}) is not limited to skills generated by FundaPod's own distillation pipeline.
Any persona pack can be loaded into the pod, whether it was distilled by other practitioners from an investor's letters and interviews, compiled from a canonical book on a strategy, or hand-curated by a domain expert, provided it declares the contract FundaPod uses to schedule it: a phase, a runner type, and the data categories it \texttt{needs} and \texttt{produces}.

The mechanism is a skills \emph{adapter}: it reads the external pack, validates the declared inputs and outputs against the data-producer skills already in the registry, and registers the pack as a Compose-phase agent skill that the planner can schedule like any internally distilled persona.
No change to the planner, the dispatcher, or any upstream Ingest or Analyze skill is required.

\paragraph{Running example: an open-source Buffett pack.}
Table~\ref{tab:bft_skill} summarises the structure of one such external pack: a publicly available encoding of Warren Buffett's investment framework at \url{https://github.com/agi-now/buffett-skills}.
The pack ships five sections (analytical modules, dispatch logic, reference content, trigger scenarios, and a fixed output format), all defined independently of FundaPod.
We use it only to make the persona-axis claim concrete; nothing in the architecture is specific to this pack.

\begin{table}[!htbp]
\centering
\caption{Structure of the open-source Buffett persona pack, as declared by its authors before onboarding.}
\label{tab:bft_skill}
\begin{tabular}{@{}p{0.22\textwidth} p{0.72\textwidth}@{}}
\toprule
\textbf{Section} & \textbf{Contents declared by the pack} \\
\midrule
Analytical modules &
\emph{Moat Scanner} (Network Effects, Efficient Scale, Switching Costs, Cost Advantage, Intangible Assets);
\emph{Margin-of-Safety Calculator} (Suggested Entry Price, Intrinsic Value Range, Optimistic / Base / Pessimistic);
\emph{Management Mirror} (Capital Allocation history, Pay--Shareholder Alignment, Promises vs.\ Delivery). \\
\addlinespace
Dispatch logic &
\emph{Quick Screen}: 8-question checklist, no reference files.
\emph{Deep Analysis}: load Moat, Financials, Valuation, Management as needed.
\emph{Topic Question}: load the single relevant reference. \\
\addlinespace
Reference content &
Eight long-form notes covering thinking frameworks, investment philosophy, business moats, management and governance, financial metrics, valuation and capital, risk and behaviour, and industry playbooks. \\
\addlinespace
Trigger scenarios &
Buy/Hold/Sell decisions; capital allocation, buybacks, dividends; industry-specific analysis; stock and company analysis; value-investing concepts; financial reports and annual letters; management-quality assessment; moat and competitive advantage; market sentiment and macro risk. \\
\addlinespace
Output format &
Final verdict (Buy / Pass / Hold / Sell); financial snapshot (3--5 line items); valuation and margin of safety; moat and management; sell-criteria check; monitoring indicators (max 3); circle of competence. \\
\bottomrule
\end{tabular}
\end{table}

\paragraph{What onboarding produces.}
After the adapter runs, the pack is materialised as a typed persona record the pod can schedule.
Table~\ref{tab:buffett_record} shows the resulting record for the Buffett pack: a stable identifier the rest of the system uses, display metadata (name, title, voice), a list of skills the persona owns, and the named workflows the dispatcher can launch on demand.
The fields are independent of the pack's provenance, so a distilled-investor skill authored by anyone else lands in the same shape.

\begin{table}[!htbp]
\centering
\caption{Internal persona record produced by the skills adapter from the external Buffett pack.}
\label{tab:buffett_record}
\begin{tabular}{@{}p{0.22\textwidth} p{0.72\textwidth}@{}}
\toprule
\textbf{Field} & \textbf{Value} \\
\midrule
\texttt{id} & \texttt{buffett} \\
\addlinespace
\texttt{name} & Warren Buffett \\
\addlinespace
\texttt{title} & Value Investor \\
\addlinespace
\texttt{sector\_hint} & Information Technology \\
\addlinespace
\texttt{voice} & Plain-spoken, folksy Omaha tone. Long-term horizon (10-year hold mindset). Skeptical of fads and complexity. Honest about what's outside the circle of competence. \\
\addlinespace
\texttt{skills} & \texttt{[buffett]} \\
\addlinespace
\texttt{default\_template} & \texttt{buffett-pitch} \\
\addlinespace
\texttt{workflows} &
\textbf{Full Pitch} (\texttt{buffett-pitch}): complete analysis of moat, management, owner earnings, intrinsic value, sell-criteria check, risks, monitoring indicators. \newline
\textbf{8-Question Filter} (\texttt{buffett-quick-filter}): fast pass through Buffett's 8-question gate (competence, durability, moat, pricing power, earnings quality, debt safety, integrity, reasonable price). \newline
\textbf{Sell Check} (\texttt{buffett-sell-check}): apply Buffett's four sell criteria (severe overvaluation, moat destruction, integrity issue, materially better opportunity). \\
\bottomrule
\end{tabular}
\end{table}

The onboarding process is the bridge: Table~\ref{tab:bft_skill} is the input the persona's author wrote, and Table~\ref{tab:buffett_record} is what the pod actually schedules.
Once a pack has been onboarded into that internal shape, it participates in every other FundaPod capability (DAG derivation, Evidence Store provenance, Knowledge Graph linkage) on equal footing with personas generated by the in-house distillation pipeline.

\section{Example outputs with and without the Buffett skill}
\label{app:examples}

This appendix is a worked instance of two of the design principles in operation.
It illustrates DP2 (\emph{independence before synthesis}, \S\ref{sec:design_principles}) by running two persona lenses on the same upstream evidence with no cross-agent context, and DP3 (\emph{provenance as a first-order requirement}) by holding the Evidence Store --- SEC filings, KPI extracts, market snapshots, and news --- identical across the two runs so that any contrast between the memos cannot be attributed to differing data access.
The first memo runs the default pitch-memo workflow with a generic persona (no Buffett pack loaded).
The second memo runs the Buffett pack from Appendix~\ref{app:buffett_skill} as the Compose-phase skill.
Only the Compose stage differs.
The systematic version of this single-target contrast, scaled to multiple companies and scored by independent finance-literate raters, is the planned human evaluation described in our future-work plan (\S\ref{sec:conclusion}); the appendix below presents the qualitative form of that contrast for one ticker, as a concrete reading aid before the formal evaluation.
We reproduce the most distinctive excerpts of each below.

\subsection{Memo A: pitch produced without the Buffett skill}
The baseline memo follows a generic pitch-memo template: a thesis paragraph, business description, recent financials, risks, catalysts, and a sources list. The thesis paragraph is reproduced below verbatim.

\begin{quote}\small
NVIDIA's Data Center segment produced \$62.3B in revenue in Q4 FY2026, 91.5\% of total sales, as Blackwell architectures became the dominant revenue driver, lifting Q4 GAAP gross margin to roughly 75.0\% and full-year revenue to \$215.9B (+65\% YoY); the Blackwell ramp represents a second hyper-growth leg, not merely an upgrade cycle. The central debate is whether hyperscaler AI capex sustains at current intensity through 2026--2027, or is disrupted by a spending pause, custom-ASIC share gains at scale, or incremental US export controls---the last of which already forced a \$4.5B H20 inventory charge and yielded only roughly \$60M in licensed China revenue in FY2026. A view change requires evidence of gross-margin compression below 70\%, a measurable hyperscaler capex deceleration, or a CUDA-ecosystem crack that materially lowers switching costs to alternative accelerator platforms.
\end{quote}

The baseline memo runs about 39 lines of markdown. It is competent and well-sourced but stylistically neutral: a roughly equal weight on each section of the pitch template, no explicit framework, and no view on whether the business sits inside any particular analyst's coverage area.

\subsection{Memo B: pitch produced with the Buffett skill}
Loading the Buffett pack does not change the data the memo is built from; it changes the reasoning structure imposed on top.
The same engagement now produces a memo organised around Buffett's analytical primitives: an 8-question quick filter, a circle-of-competence assessment, an explicit list of key assumptions with stated falsification conditions, a moat analysis with widening and narrowing signals, a management-and-capital-allocation section, a financial snapshot with ROIC and cash-conversion computed inline, and a final buy/don't-buy conclusion with an explicit price at which the view would change.
The memo runs about 389 lines, roughly ten times the length of the baseline, because the framework forces explicit reasoning at each step.

A representative excerpt is the circle-of-competence assessment, which the baseline memo has no analogue for:

\begin{quote}\small
\textbf{Boundary area --- leaning outside.}
I can describe NVIDIA's business in a paragraph: it designs the world's best parallel-computing chips and, more importantly, the CUDA software platform that locks developers into its ecosystem.
What I \emph{cannot} do with confidence is forecast whether CUDA's grip endures across multiple architecture generations while hyperscalers deploy hundreds of billions in custom silicon, inference-time compute efficiency improves via algorithmic advances, and geopolitics reshapes the global semiconductor supply chain. The technological substrate here is changing faster than anything in my six decades of investing. \ldots\ I will nonetheless complete the analysis because the financial quality is so extraordinary and the market's pricing of this name deserves a careful examination.
\end{quote}

The final conclusion paragraph is similarly stylised:

\begin{quote}\small
\textbf{Don't Buy.} NVIDIA is a genuinely exceptional business---the closest thing to a franchise in AI infrastructure---with extraordinary cash generation, a founder-CEO of singular vision, and a fortress balance sheet. But at \$5.70 trillion market cap (\$235/share) the stock prices in flawless execution across a decade of rapid technological change, leaving no margin of safety for the inevitable unknowns: export-control escalation, hyperscaler custom ASICs, or AI compute-efficiency breakthroughs. The business also sits at the outer edge of my circle of competence; I cannot reliably estimate ten-year owner earnings when the underlying technology architecture could shift materially within five. The price that would compel serious reconsideration is roughly \$77--\$115/share---less than half today's price.
\end{quote}

\subsection{What changes and what does not}
The two memos see the same Evidence Store artifacts (10-K segments, KPI tables, market snapshots, news). What changes is which questions the Compose-phase agent asks, in what order, and against what reference rubric. The Buffett memo's distinctive features (the 8-question filter, the circle-of-competence disclaimer, the ROIC and cash-conversion computations, the explicit price-trigger reconsideration band, and the management-integrity audit) all come from the persona pack rather than from any data the baseline memo did not have access to.
This is independence-then-synthesis (DP2, \S\ref{sec:design_principles}) made concrete in a single engagement: the two lenses run in isolation under a shared provenance contract (DP3); the contrast is not collapsed into a consensus output but surfaced for the PM to inspect and adjudicate. The single-instance qualitative comparison shown here is the small-scale precursor to the planned multi-company, blind-rated evaluation outlined in \S\ref{sec:conclusion}, which formalizes the same comparison structure across a pre-registered sample.

\end{document}